\pdfoutput=1
\PassOptionsToPackage{table,dvipsnames}{xcolor}

\documentclass[11pt]{article}

\usepackage[preprint]{acl}

\usepackage{times}
\usepackage{latexsym}

\usepackage[T1]{fontenc}

\usepackage[utf8]{inputenc}

\usepackage{microtype}

\usepackage{inconsolata}
\usepackage{xcolor}

\usepackage{graphicx}
\usepackage{amssymb}
\usepackage{tabularx}
\usepackage{makecell}
\usepackage{booktabs}
\usepackage{multirow}
\usepackage{lscape}
\usepackage{pbox}
\usepackage{amsmath}
\usepackage{tcolorbox}
\usepackage{collcell}
\usepackage{pgf}
\usepackage{longtable}
\usepackage{caption}
\usepackage{subcaption}

\usepackage{float}

\definecolor{ColorBlue}{RGB}{0, 114, 178}
\definecolor{ColorOrange}{RGB}{230, 159, 0}
\definecolor{ColorTurquoise}{RGB}{86, 180, 233}
\definecolor{ColorRed}{RGB}{213, 94, 0}
\definecolor{ColorPurple}{RGB}{204, 121, 167}
\definecolor{ColorGreen}{RGB}{0, 158, 115}

\title{Sightation Counts: Leveraging Sighted User Feedback \\ in Building a BLV-aligned Dataset of Diagram Descriptions}
\author{
    \textbf{Wan Ju Kang}\textsuperscript{$\alpha$} \quad
    \textbf{Eunki Kim}\textsuperscript{$\alpha$} \quad
    \textbf{Na Min An}\textsuperscript{$\alpha$} \quad
    \textbf{Sangryul Kim}\textsuperscript{$\alpha$} \\
    \textbf{Haemin Choi}\textsuperscript{$\beta$,$\delta$} \quad
    \textbf{Ki Hoon Kwak}\textsuperscript{$\gamma$,$\delta$} \quad
    \textbf{James Thorne}\textsuperscript{$\alpha$} \\
    KAIST AI\textsuperscript{$\alpha$}\quad
    Sungkyunkwan University\textsuperscript{$\beta$} \quad
    Yonsei University\textsuperscript{$\gamma$}\\
    Work done as KAIST AI research intern\textsuperscript{$\delta$}\\
    \textsuperscript{$\alpha$}\small{\texttt{\{soarhigh, eunkikim, naminan, sangryul, thorne\}@kaist.ac.kr}} \\
    \textsuperscript{$\beta$}\small{\texttt{chm1009@g.skku.edu}} \quad \textsuperscript{$\gamma$}\small{\texttt{kihoon090@yonsei.ac.kr}}\\
    \raisebox{-0.3\height}{\includegraphics[height=14pt]{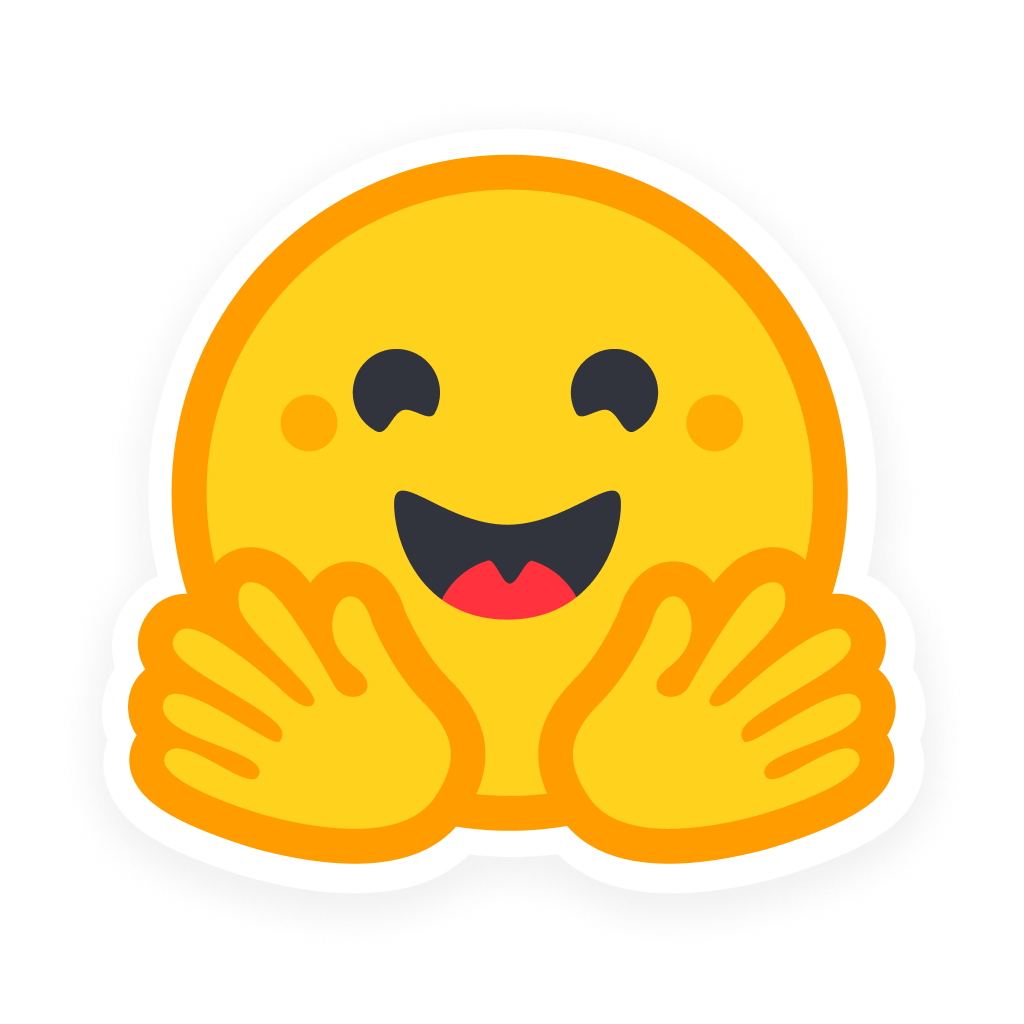}}~\small{\url{https://hf.co/Sightation}}
}

\begin{document}

\maketitle

\begin{abstract}
Often, the needs and visual abilities differ between the annotator group and the end user group. Generating detailed diagram descriptions for blind and low-vision (BLV) users is one such challenging domain. Sighted annotators could describe visuals with ease, but existing studies have shown that direct generations by them are costly, bias-prone, and somewhat lacking by BLV standards. In this study, we ask sighted individuals to assess---rather than produce---diagram descriptions generated by vision-language models (VLM) that have been guided with latent supervision via a multi-pass inference. The sighted assessments prove effective and useful to professional educators who are themselves BLV and teach visually impaired learners. We release \textsc{Sightation}, a collection of diagram description datasets spanning 5k diagrams and 137k samples for completion, preference, retrieval, question answering, and reasoning training purposes and demonstrate their fine-tuning potential in various downstream tasks\footnote{Wherever possible, we use color blind safe palettes in figures and tables.}. 

\end{abstract}

\section{Introduction}

\begin{table*}[h]
    \centering
    \small
    \definecolor{blue}{RGB}{16,85,154}
    \definecolor{pink}{RGB}{219, 76, 119}
    \newcommand{\checkblue}{\textcolor{blue}{\checkmark}}
    \newcommand{\checkpink}{\textcolor{pink}{\texttimes}}
    
    \begin{tabularx}{\textwidth}{p{4.2cm} p{1.7cm} p{1.6cm} p{2.2cm} X}
        \toprule
        \makecell[c]{\textbf{Dataset}} & \textbf{\makecell[c]{Average \\Text \\Length}} & \textbf{\makecell[c]{Validated \\by\\ BLV?}} & \makecell[c]{\textbf{Applications}} & \textbf{\makecell[c]{Dimensions\\Assessed}}\\
        \midrule
        \makecell[l]{\textbf{\textsc{Sightation} (Ours)}\\
            \hspace{0.6cm}\textsc{-Completions}\\
            \hspace{0.6cm}\textsc{-Preference}\\
            \hspace{0.6cm}\textsc{-Retrieval}\\
            \hspace{0.6cm}\textsc{-VQA}\\
            \hspace{0.6cm}\textsc{-Reasoning}
        } & \makecell[r]{\textbf{188.3}\\(words)} & \makecell[c]{\textbf{\checkblue}} & \makecell[l]
        {$\cdot$ \textbf{Completion} \\
        $\cdot$ \textbf{Preference}\\
        ~~~\textbf{alignment} \\
        $\cdot$ \textbf{Retrieval}\\
        $\cdot$ \textbf{Reward} \\
        ~~~\textbf{modeling} \\
        $\cdot$ \textbf{Question} \\
        ~~~\textbf{answering}
        } & \makecell[l]{$\cdot$ \textbf{Factuality} \\ $\cdot$ \textbf{Informativeness} \\ $\cdot$ \textbf{Succinctness} \\ $\cdot$ \textbf{Diversity} \\ $\cdot$ \textbf{Usefulness}, \\~~~in 4 finer aspects \\ $\cdot$ \textbf{Interpretiveness} \\ $\cdot$ \textbf{Preferred Description} \\ $\cdot$ \textbf{Best Sentence}} \\
        \midrule
        VisText \cite{tang2023vistext} & \makecell[r]{74.6} & \makecell[c]{\checkpink} & Completion & Accuracy, Descriptiveness \\
        MathVista \cite{lu2023mathvista} & \makecell[r]{58.0} & \makecell[c]{\checkpink} & VQA, Reasoning & Correctness \\
        ChartGemma \cite{masry2024chartgemma} & \makecell[r]{37.5} & \makecell[c]{\checkpink} & Completion & Informativeness, Factual Correctness, Structure \\
        DiagramQG \cite{zhang2024diagramqg} & \makecell[r]{9.5} & \makecell[c]{\checkpink} & DQA & Diversity, Object Density \\
        VizWiz-VQA \cite{gurari2018vizwiz} & \makecell[r]{8.6} &  \makecell[c]{\checkblue} & VQA & Diversity, Answerability \\
        VizWiz-LF \cite{huh2024long} & \makecell[r]{73.2} & \makecell[c]{\checkblue} & VQA & Relevance, Helpfulness, Plausibility, Fluency, Correctness \\
        \bottomrule
    \end{tabularx}
    \caption{The \textsc{Sightation} collection has been validated by teaching professionals who are visually impaired and are experienced instructors at schools for the blind. As the most text-dense diagram description dataset to date, it can be used to drive a variety of training objectives towards BLV accessibility needs. We discuss a few prime examples in Section~\ref{sec:perf}. This table includes only the few most closely related works; we deliver an extended comparison in Table~\ref{tab:dataset_comparison_extended}.}
    \label{tab:dataset_comparison}
\end{table*}

Recent research has seen rapid development in vision-language models (VLM). Seeing the world and the data within has significantly advanced machine intelligence in a variety of tasks \cite{Liu_2024_CVPR, zhu2023minigpt, yang2024qwen2, qwen2025qwen25technicalreport, xu2024lvlm, li2024temporal}, reaching a fast-growing user pool with quicker and easier access.

However, the same cannot be said of blind and low-vision (BLV) individuals. Widely adopted evaluation metrics have been shown to be biased against their preferences \citep{kapur-kreiss-2024-reference} and benchmark studies tend to pursue a larger audience first \citep{li2024seed, li2024vlrewardbench}. Publicly available reward models for generic VLMs are scarce \citep{zang2025internlm} --- let alone for the visually impaired. Vision-language dataset research appears divided between breadth \citep{tang2023vistext, lu2023mathvista}, specificity \citep{masry2024chartgemma, masry-etal-2024-chartinstruct}, and volume \citep{zhang20252, lee2022multimodal}.

Perhaps the classroom setting best exemplifies the circumstances BLV individuals face: textual information is combined with images (such as diagrams, graphs, and figures) to help learners fully grasp complex information \citep{vekiri2002value, cheng2009towards, tippett2016recent, gates2018importance}. VLMs at the command of BLV users must therefore provide select, curated information rather than an indiscriminate narration of data.

Instilling this behavior in VLMs, however, remains challenging primarily due to dataset concerns. The unavailability of large-scale BLV-aligned datasets has prompted previous studies to crowdsource a few expert sighted annotators to \textit{generate} descriptions. The limitation of this approach is twofold: \textit{\romannumeral 1)} it does not account for the preference misalignment between the BLV evaluator and the sighted generator \citep{9555469}; \textit{\romannumeral 2)} it is prone to modeling the generations after the annotator rather than the task, introducing annotator bias into the dataset \citep{geva2019we}.  While \citet{kreiss2022context} has illustrated the potential of sighted users as BLV preference estimators for a few specific qualities of generations, whether their findings will generalize to a dataset-scale volume of generations or with other aspects of perceived quality remains unknown. 

We construct, what is to the best of our knowledge, the first dataset that addresses the union of aforementioned challenges. We prompt a VLM to generate a guide, which will be input to a second inference pass to latently supervise the second-pass behavior in favor of BLV users. Then, we further invoke the VLM to generate diagram descriptions, saving on crowdsourcing cost and reducing annotator fatigue. We distribute to sighted annotators a set of assessment tasks, substantially less demanding than a generation task, implying easier recruiting of a sufficiently large annotator population, potentially mitigating annotator bias. Finally, we design the assessment tasks such that they are finer-grained than any prior work we are aware of.

The compilation we named \textsc{Sightation} is the first large-scale BLV-aligned dataset that is validated by BLV professionals and can be used to train on a broad range of objectives. A few statistics to highlight our dataset performance include: preference-tuning a 2B model on our dataset to achieve an average $1.67\sigma$ increase in the usefulness rated by the BLV group; instruction-tuning a 2B model on our dataset to outperform a 3B model fine-tuned on chart comprehension \citep{masry2024chartgemma} in 8 out of 11 automatic metrics; contrastive tuning a \textsc{BLIP-2}\citep{li2023blip} for retrieval purposes to outperform a COCO-tuned \textsc{BLIP-2} by 65\%p on Precision@1.

\begin{figure*}[t]
    \centering
    \includegraphics[width=1.0\linewidth]{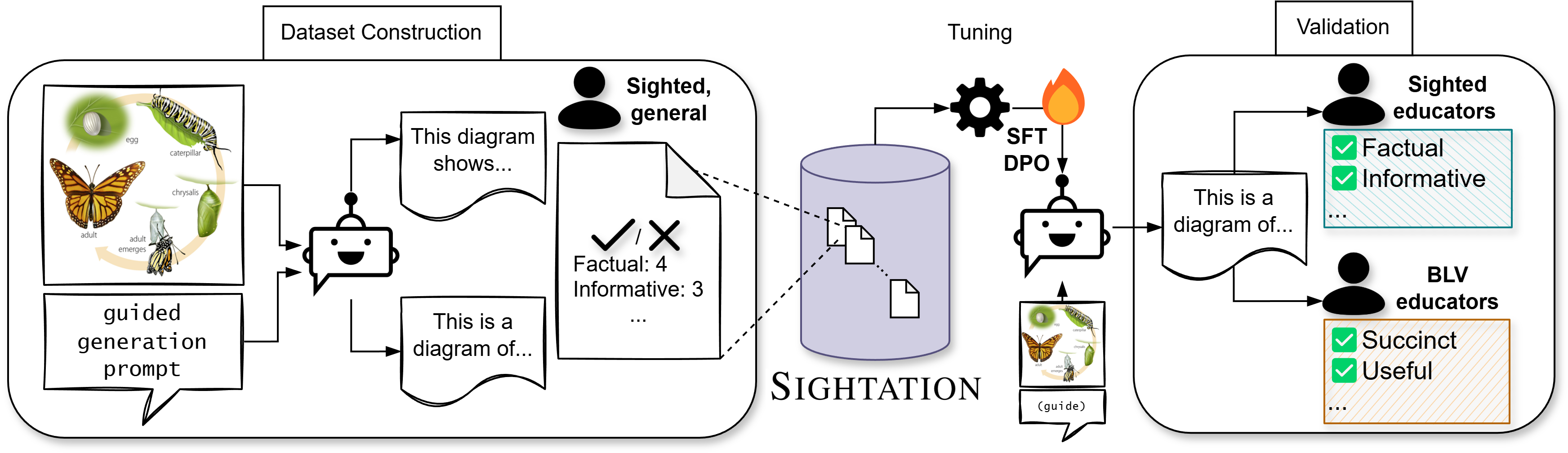}
    \caption{The key benefit of utilizing sighted user feedback lies in their \textit{assessments}, which are based on solid visual grounding. The compiled assessments prove an effective training substance for steering VLMs towards more accessible descriptions. Dataset use and the subsequent validation are described in Sec.~\ref{sec:perf}. A complete list of use cases is provided in Appendix~\ref{app:rest}.}
    \label{fig:main_visual}
\end{figure*}
\begin{figure}[b]
    \centering
    \includegraphics[width=0.95\linewidth]{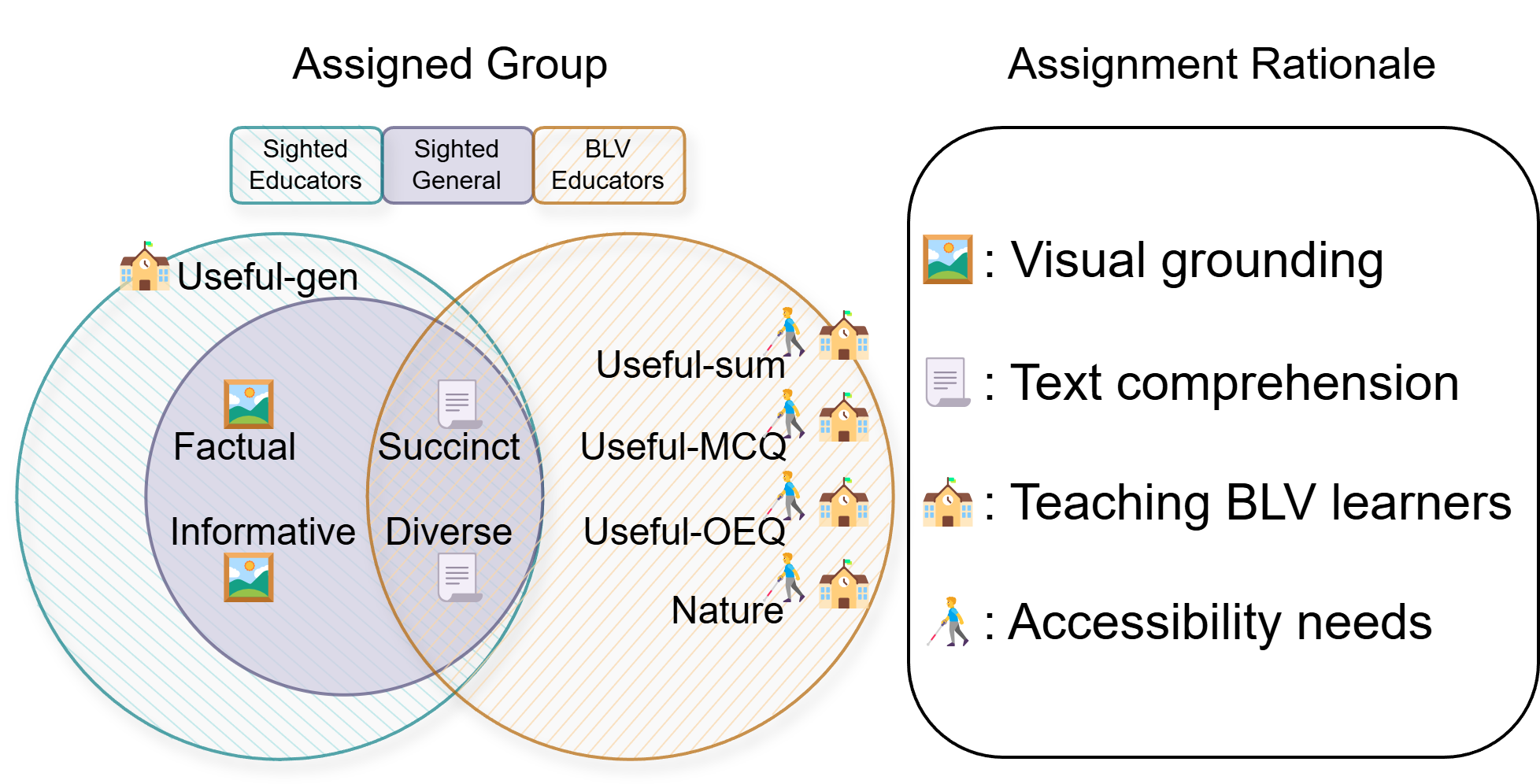}
    \caption{The qualities assessed by their respective groups.}
    \label{fig:dimension_assignments}
\end{figure}

\section{Related Work}

\textbf{Accessibility Studies.} \citet{9555469} found that BLV and sighted reader groups differ significantly on which semantic content they consider as most useful, suggesting that access to meaningful information is strongly reader-specific. VizWiz-VQA \cite{gurari2018vizwiz} contains images and visual QA pairs produced by blind people encouraging the development of more generalized algorithms that can assist the blind. As an extended work, VizWiz-LF \cite{huh2024long} includes long-form answers from BLV people. VisText \cite{tang2023vistext} contains charts and captions that convey different levels of semantic content. As shown in Table~\ref{tab:dataset_comparison}, VizWiz-VQA and VizWiz-LF were validated by BLV users but only focus on Visual QA (VQA) applications. VisText examines the role of the level of semantic content but was not validated by BLV for dataset purposes. As a diagram description dataset validated by BLV users, \textsc{Sightation} explores diverse use cases, with assessments on various aspects. 

\noindent \textbf{Image Description Tasks and Models.} \citet{wang2024qwen2} presented the \textsc{Qwen2-VL} collection, which includes three open-weights models: 2B, 7B, and 72B. \textsc{Qwen2-VL} matches the performance of \textsc{GPT-4o} and \textsc{Claude3.5-Sonnet} \cite{anthropic2024claude} in multimodal scenarios, surpassing other open-weights VLMs at the time. 

\textsc{GPT-4o} \cite{hurst2024gpt} accepts multimodal input and generates high-quality outputs including text and codes, showing powerful multimodal understanding capability. Using these VLMs, the image description task aims to generate a descriptive textual context for images of different types (\textit{e.g.}, photographs, illustrations, schematics, and diagrams). Flickr8K and PASCAL-50S comprise natural images, captions, and human judgments\cite{hodosh2013framing, vedantam2015cider}, and Polaris \cite{wada2024polos} incorporated synthetic captions from image captioning models.

ChartGemma \cite{masry2024chartgemma} contains chart images collected from specialized websites and instruction-tuning data generated from the charts. MathVista \cite{lu2023mathvista} encompasses diverse visual contexts from natural images to diagrams or plots that require mathematical reasoning. However, Table~\ref{tab:dataset_comparison} shows that these datasets have an average text length much shorter than ours, even though charts and mathematical images could be highly information-dense. Complementing the limitation, \textsc{Sightation} provides contexts that top in average text length to date with variants for downstream tasks.

\noindent \textbf{Human Annotation Efforts.} 
Human judgment annotations are essential in evaluating image captions, complementary to automatic metrics. Common approaches involve employing annotators to assess captions based on rating scales for specific dimensions of text quality \cite{gehrmann2023repairing}. However, it comes with challenges, including subjectivity and consistency issues. \citet{amidei-etal-2019-agreement} argues that the evaluation of generated text is intrinsically subjective and relies on different factors including annotator experience, motivation, knowledge, or education.  A related line of research \citep{glockner-etal-2024-ambifc, nie2020can} directly addressing this limitation advocates that generations from few-annotator pools fall short in terms of coverage of the distribution of opinions.

\section{The \textsc{Sightation} Dataset}\label{sec:sightation}

\textsc{Sightation} is a BLV-specific vision-language dataset for the educational domain. It is built upon the AI2D dataset \citep{kembhavi2016diagram}: we chose this for two reasons: it contains diagrams from grade school material, requiring no specialized expertise or domain knowledge in our annotator recruiting process; diagrams pose a unique challenge to VLMs in that they often require an understanding of the rendered schematics \textit{and} the natural objects.

AI2D contains 5k science diagrams, with 150k annotations, spanning OCR texts and bounding box locations, as well as 15k multiple choice questions. Of these features, we take only the diagrams, to simplify \textsc{Sightation}-like dataset construction in the future. All notation and labeling methods used in this section are summarized in a separate Table~\ref{tab:notation} to aid comprehension.

\subsection{Overview}

Different annotator roles can be found in Figure~\ref{fig:dimension_assignments}. There are a total of 9 aspects to be assessed, and these were inspired by various related studies. In \citet{kreiss2023contextref}, relevance and irrelevance aspects are studied to measure the image information carried in text and the inclusion of extraneous information in the text, respectively. As such, we chose to examine \textbf{Informativeness} and \textbf{Factuality} dimensions. These both require reliable visual grounding so were assigned to the sighted accordingly. We also opted for some measures to be assessed by all groups. Since brevity \citep{9555469} and diverse opinion coverage \citep{glockner-etal-2024-ambifc, nie2020can} have been pointed out as contributors to perceived quality, we chose to incorporate them as the \textbf{Succinctness} and the \textbf{Diversity} aspects, both of which are assessable with text comprehension alone. Following \citet{tang2023vistext}, we split the use cases for the usefulness measure along typical vision-language comprehension tasks common in the classroom: \textbf{Useful-Sum} (summarization), \textbf{Useful-MCQ} (multiple-choice questions), and \textbf{Useful-OEQ} (open-ended questions). These were assigned to the BLV educators, adept at teaching and knowledgeable in accessibility needs. A general usefulness measure \textbf{Useful-Gen} was assigned to the sighted educators to probe their estimate of BLV needs. Finally, a categorical variable, \textbf{Nature}, was assigned to the BLV educators to ask for their opinion on how interpretive the text appears.

These different subsets were assigned to pursue a synergistic interplay between varying visual abilities, teaching experience, and accessibility requirements. The sighted general group, shown on the left in Figure~\ref{fig:main_visual} ensures that the diagram \textit{content} is well-conveyed in the description. Sighted educators, shown on the top right of Figure~\ref{fig:main_visual} validate the general group's assessment whilst also rating the general usefulness of the description to BLV users. Finally, the text-based assessment by BLV educators, shown on the bottom right in the same figure, gauges the alignment of \textsc{Sightation}-tuned descriptions with BLV preferences. A more detailed description of the annotation tasks is in Section~\ref{sec:annot} for the sighted general group and in Section~\ref{sec:pros} for the sighted and BLV.

\subsection{Guided Generation with Latent Supervision}\label{sec:guide}

Previous work \citep{9555469} has shown that crowdsourced data visualization descriptions written by sighted crowdworkers were not equally useful to the BLV groups as they were to the sighted, in terms of describing low-level numerical elements or high-level insights such as subjective commentary. Building on this, we hypothesized that the key to generating a description that is useful to BLV individuals lies not only in \textit{what} is seen but also in \textit{how} the perceived information is articulated. We hypothesized that introducing auxiliary data such as plausible question-answer pairs, would have a good effect as they assist the description generator with understanding which parts are critical and which are less so.

In implementing this idea, we incorporated a two-pass guided generation process. The first inference pass is to create the guide, which is a VLM-generated set of question-answer pairs in response to an input diagram. We carefully examine the quality of the question and answer pairs we have generated and, in the Appendix \ref{app:gqa}, provide a more in-depth analysis of how these pairs differ from those originally included in the AI2D dataset. Then, the second pass generates the diagram description in response to the input diagram \textit{and} the guided generation prompt, as shown on the leftmost part of Figure~\ref{fig:main_visual}. 

We applied this generation process with two models: \textsc{GPT-4o mini} and \textsc{Qwen2-VL} 72B model, producing four descriptions for each of the 5k diagrams in the AI2D dataset. The working dataset thus contains 20k descriptions.

\subsection{Annotation Tasks}\label{sec:annot}

1k images were randomly sampled from the working dataset. They were then paired with their respective descriptions generated by \textsc{GPT-4o mini} ($\textbf{Desc}_{}^{\texttt{g}}$ and $\textbf{Desc}_{\texttt{++}}^{\texttt{g}}$) and descriptions generated by \textsc{Qwen2-VL} ($\textbf{Desc}_{}^{\texttt{q}}$ and $\textbf{Desc}_{\texttt{++}}^{\texttt{q}}$) were distributed to the 30 sighted annotators, to complete three tasks: \textit{\romannumeral 1)} preference choice, \textit{\romannumeral 2)} quality rating, and \textit{\romannumeral 3)} best sentence choice. The 1k tuples were partitioned into 10, so that 3 participants perform the annotation on a shared total of 100 tuples.

First, annotators were asked to select pairwise preferred descriptions: one from the GPT pair and the other from the Qwen pair. Second, for all four diagram descriptions, they were asked to rate the description quality across the 4 aspects assigned to them, as in Figure~\ref{fig:dimension_assignments}, on a 5-point Likert scale.

Lastly, they were asked to pick the best-contributing sentence from each of the four diagram descriptions. Sample screenshots of the annotation interface, along with the annotation guidelines, are provided in Appendix~\ref{app:guidelines}.

The total number of annotations is 11,804, spanning 998 diagrams and 3,992 descriptions. Further statistics and post-processing steps are found in Appendix~\ref{app:ann_detail}.

\subsection{Dataset Construction}
In this section, we describe how the annotated tuples are processed for various downstream tasks.

\subsubsection{Chat Completion} 

\textsc{SightationCompletions} contains instruction-response pairs from two sets: \textit{\romannumeral 1)} all the 4k human-annotated descriptions over 1k images, with the base instruction in Appendix~\ref{app:prompts} and \textit{\romannumeral 2)} the top 25\% highly rated descriptions for each of the 4 aspects annotated. For the latter subset, we augment the base instruction to pair responses that were of high quality in some aspect. We append an aspect-specific suffix outlining the desired quality according to our annotation guidelines in Appendix~\ref{app:guidelines}. For instance, the aspect suffix for the factuality dimension is: ``When generating the diagram description, pay close attention to making it factual. A highly factual description delivers only the facts that are grounded in the diagram.''

With the former set consisting of 4k (diagram, base prompt, description) samples and the latter set consisting of 1k (diagram, augmented prompt, description) samples per aspect, our completions dataset totals 8k samples. 

\subsubsection{Preference Alignment} 

\textsc{SightationPreference} also proceeds from the 4k diagram-description pairs, consisting of 4 descriptions for every image. From these 4, we take the 6 possible pairwise combinations and label ``chosen'' and ``rejected'' to each contender in the pairwise comparisons as follows.
\paragraph{In-model Contenders}
Within each of the 2 same-model comparisons, (\textit{e.g.}, $\textbf{Desc}^{\texttt{g}}_{}$ versus $\textbf{Desc}^{\texttt{g}}_{\texttt{++}}$) we directly take the $\textbf{Preference}^{model}$ annotation to assign ``chosen'' and ``rejected''. This assignment results in 2 \texttimes~1k = 2k chosen-rejected preference pairs.
\paragraph{Cross-model Contenders}
Within each of the 4 cross-model comparisons, (\textit{e.g.}, $\textbf{Desc}^{\texttt{g}}_{\texttt{++}}$ versus $\textbf{Desc}^{\texttt{q}}_{}$), we averaged the rating scores per contender and assigned\footnote{Ties are technically possible, but the collected annotations did not contain any.}  ``chosen'' to the ratings winner. This assignment results in 4 \texttimes~1k = 4k preference pairs.
\paragraph{Synthetic Contenders}
Additionally, we synthesized an inferior (``rejected'') variant of a description by removing its best sentence. To account for the reduced length, we remove a random non-best sentence from the original description and label this variant ``chosen''. This assignment results in 4 \texttimes~1k = 4k preference pairs per annotator. A maximum of three annotators evaluated the same sample, so the preference pairs total 12k. After deduplicating (\textit{e.g.}, annotators selecting the same sentence as the best sentence), we have 10k preference pairs.

Putting together the in-model (2k), cross-model (4k), and synthetic (10k) contenders and their respective labels, \textsc{SightationPreference} spans 16k pairs.

\subsubsection{Retrieval}         

Each row in \textsc{SightationRetrieval} contains an image as a retrieval query, accompanied by the top 1, top 5, and top 10 descriptions as the positives, as well as 10 hard negatives. This set contains 1k rows, with a potential well beyond that number. For instance, more than 63 million unique combinations can be derived utilizing 5 random samples from the 10 positives and 5 random samples from the 10 negatives. Further details can be found in Appendix~\ref{app:retrieval}.

\section{Performance Analysis}\label{sec:perf}

We designed a series of experiments to measure the performance of \textsc{Sightation} as a \textit{dataset}. First, we fine-tuned various models on our dataset. Then, we asked sighted and BLV teachers at schools for the blind to evaluate the generated texts. Additionally, we employ VLM judges and a number of well-known classic metrics to evaluate the descriptions. We report the main findings on the extent and breadth of performance enhancement our dataset can cultivate.

\subsection{Fine Tuning}

We chose to experiment with the \textsc{Qwen2-VL} series \citep{wang2024qwen2} considering its size variety, state-of-the-art performance at the time of writing, as well as whether the largest variant (72B) could fit on our compute cluster in its default precision, \texttt{bf16}, unquantized. We fine-tuned the 2B and 7B models and performed comparative analyses. Finer details on the tuning configuration are found in Appendix~\ref{app:tuning_config}.

\subsubsection{On \textsc{SightationCompletions}}

We conducted supervised fine tuning (SFT) on our completions dataset. The 2B model underwent full fine tuning, whereas the 7B model underwent parameter-efficient fine tuning (PEFT).

\subsubsection{On \textsc{SightationPreference}}

For preference alignment tuning, we chose to perform Direct Preference Optimization (DPO, \cite{rafailov2024direct}). Since reward models trained on generic data may not accurately represent BLV preferences, we opted for DPO, a widely used algorithm free of reward models. Before the actual DPO training, as is common in practice, we first subjected the 2B and 7B models to SFT. However, we recognized that sharing the same set of diagrams across the SFT and DPO stages could pose higher overfitting risks. With that in mind, instead of using \textsc{SightationCompletions} for SFT, we randomly sampled 1k diagrams along with their 4 descriptions from the remaining pool of generated descriptions (\textit{i.e.}, the ones not in \textsc{SightationCompletions}) and used these to compile 4k completion samples. Afterwards, DPO was run on \textsc{SightationPreference}. At both the SFT and DPO stages, the 2B model was fully fine-tuned, and the 7B model was trained with PEFT.

\subsubsection{On \textsc{SightationRetrieval}}

We performed contrastive training to fine-tune \textsc{BLIP-2}~\citep{li2023blip} for its appeal in image-text matching. To save compute, we trained only parts of the model and with just the top 1 positive and a randomly chosen negative. The training was carried out with InfoNCE loss~\citep{oord2018representation}, a widely used choice for contrastive objectives.

\subsection{Evaluation Setup}

\subsubsection{By Teaching Professionals}\label{sec:pros}

We recruited 17 specialized educators who teach BLV learners at schools for the visually impaired. 8 of them are themselves blind or have low vision; remaining 9 are sighted. We refer to these groups as the BLV educator group and the sighted educator group, respectively. Their demographics are reported in Tables~\ref{tab:blv_educators_demo} and ~\ref{tab:sighted_educators_demo}

\begin{figure*}[h!]
    \centering
    \includegraphics[width=0.95\linewidth]{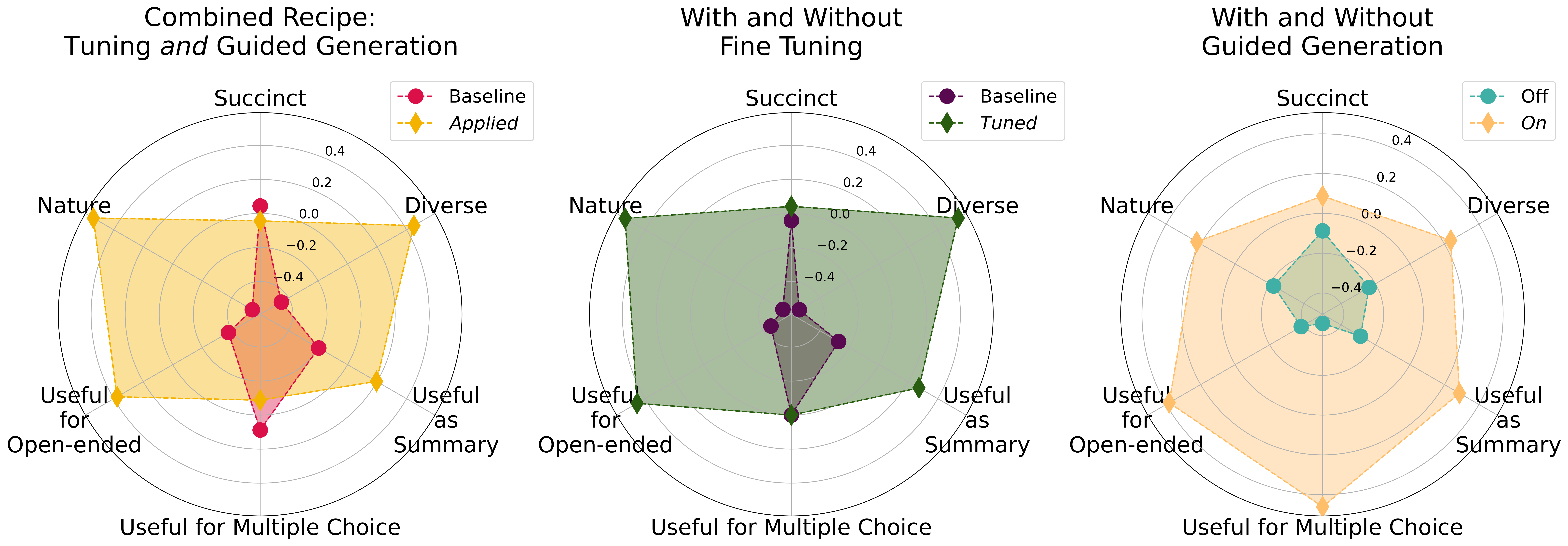}
    \caption{Tuning VLMs on \textsc{Sightation} enhanced various qualities of the diagram descriptions, evaluated by BLV educators, and shown here as normalized ratings averaged in each aspect. The capability of the dataset is most strongly pronounced with the 2B variant, shown above. Full results across 4 models and 22 metrics are reported in Tables~\ref{tab:big1a}, ~\ref{tab:big1b}, ~\ref{tab:big2B}, and ~\ref{tab:big7B}.}
    \label{fig:blv_ratings}
\end{figure*}

\paragraph{BLV Educators}

Each BLV educator was given 40 diagrams, each with two competing descriptions. They were asked to rate text-based qualities. They were asked to perform a quantitative assessment on the aspect set pictured in Figure~\ref{fig:dimension_assignments}.

Following \citet{tang2023vistext,9555469}, we chose to investigate the usefulness of the diagram descriptions, but in three finer manifestations. Specifically, we asked the BLV educators to assess how useful the description is as a textual aid providing \textit{\romannumeral 1)} a summary of the diagram content, \textit{\romannumeral 2)} clues that would be helpful when solving short-answer multiple-choice questions about the diagram, and \textit{\romannumeral 3)} clues that would be helpful when answering long-answer open-ended questions about the diagram.

\paragraph{Sighted Educators}

Each sighted educator was given 40 diagrams, each with two competing descriptions with randomized order of presentation. They were then asked to evaluate the descriptions according to the guidelines for the sighted educator group, found in Appendix~\ref{app:guidelines}. Their aspect set, also shown in Fig.~\ref{fig:dimension_assignments}, includes a usefulness estimate to BLV users.

\subsubsection{By Automatic Metrics}
We perform a VLM-as-a-Judge (\citet{dubois2023alpacafarm},\citet{zheng2023judgingllm}) evaluation with \textsc{QVQ-72B-Preview}, where we instruct the VLM to take the \textbf{Image}, $\textbf{Desc}^{model}_{}$, and $\textbf{Desc}^{model}_{\texttt{++}}$ triplet as input and produce a JSON-formatted evaluation with the same aspects as with the human annotation.

As for classic metrics, we collect widely recognized reference-free metrics since the AI2D dataset does not contain references: CLIP score \citep{hessel2021clipscore}, SigLIP score \citep{zhai2023sigmoid}, BLIP-2 Retrieval score \citep{li2023blip}, Self-BLEU (based on BLEU \citep{papineni2002bleu}), PAC score \citep{sarto2023positive}, and LongCLIP-B/L \citep{zhang2024longclip}. For the retrieval task, we chose to measure recall@$K$ and precision@$K$ for $K=1, 5, 10$, as do numerous retrieval studies.

\section{Results}\label{sec:november}

We report the evaluation results by the BLV educator group, the sighted educator group, VLM judges, and classic metrics. For each group, we discuss the effectiveness of the combined recipe, then with the guided generation ablated, and with the tuning step ablated. Here, we focus on the evaluation by BLV; sighted educator and VLM-as-a-Judge evaluation, as well as classic metric results are found in Appendix~\ref{app:results}.

\subsection{Evaluation by BLV Educators}

Here, we conduct an analysis of effect size, an intuitive choice for aggregate analysis on different sample sets rated by different evaluators. Figure~\ref{fig:blv_ratings} shows the effect size computed from BLV educators' assessment. The radial axis corresponds to the mean ratings on each of the two sets of samples under comparison, normalized by their pooled standard deviation ($\sigma$). Naturally, the radial axis is in units of the pooled standard deviation.

The first radar chart in Figure~\ref{fig:blv_ratings} shows the result of comparing $\textbf{Desc}^\texttt{q2bbase}$ and $\textbf{Desc}^\texttt{q2bdpo}_\texttt{++}$. The latter was rated more than $1\sigma$ higher in interpretiveness (\textbf{Nature}); $0.8\sigma$ better in diversity and usefulness for open-ended questions; $0.4\sigma$ units more useful as a summary.

In the middle of the same figure is shown the ablated result of fine tuning, with the guided generation turned on for both sets: a comparison between $\textbf{Desc}^\texttt{q2bbase}_\texttt{++}$ and $\textbf{Desc}^\texttt{q2bdpo}_\texttt{++}$. All 6 aspects were judged in favor of the latter, with as large as $1.2\sigma$ difference in interpretiveness and diversity and $0.8\sigma$ in usefulness for open-ended questions.

On the right is shown the effect of the guided generation on a \textsc{SightationPreference}-tuned 2B model: a comparison between $\textbf{Desc}^\texttt{q2bdpo}$ and $\textbf{Desc}^\texttt{q2bdpo}_\texttt{++}$. Guided generation yields significant enhancement for the DPO-tuned case, with $1\sigma$ higher in usefulness for multiple choice questions, followed by approximately $0.8\sigma$ improvement in usefulness for open-ended questions, an overall improvement in every aspect down to succinctness, with $0.2\sigma$. However, as will be discussed with Table~\ref{tab:blv_gg}, this effect by the guided generation is achieved only after the model is fine-tuned on our dataset, implying that a good alignment is a pre-requisite for attempting to benefit from test-time prompting.

\begin{table*}[h]
    \centering
    \small
    \begin{minipage}[t]{0.25\textwidth}
        \centering
        \tiny
        \begin{tabular}{l>{\columncolor{gray!15}}rr}

            \toprule
            & \multicolumn{2}{c}{\textbf{Combined Effect Size}} \\\cmidrule{2-3}
            \textbf{Aspect} & 2B & 7B \\ \midrule
            Succinct & -0.09 & 1.69 \\
            Diverse & 0.90 & 0.46 \\
            Useful-Sum & 0.39 & 0.53 \\
            Useful-MCQ & -0.18 & 0.20 \\
            Useful-OEQ & 0.76 & 0.00 \\\cmidrule{1-3}
            Average & 0.36 & 0.58 \\
            Nature & 1.08 & -2.38 \\
            \bottomrule
        \end{tabular}
        \caption{Combined recipe effect size on each aspect, measured with BLV assessment.}
        \label{tab:blv_cr}
    \end{minipage}%
    \hfill
    \begin{minipage}[t]{0.40\textwidth}
        \centering
        \tiny
        \begin{tabular}{l>{\columncolor{gray!15}}r>{\columncolor{gray!15}}rrr}
            \toprule
            & \multicolumn{4}{c}{\textbf{Tuning Effect Size}} \\\cmidrule{2-5}
            \textbf{Aspect} & 2B & 2B+GG & 7B & 7B+GG \\ \midrule
            Succinct & 0.06 & 0.08 & 0.37 & -0.11 \\
            Diverse & 0.87 & 1.08 & -0.06 & 0.00 \\
            Useful-Sum & 0.20 & 0.55 & 0.14 & 0.36 \\
            Useful-MCQ & 0.29 & 0.00 & -0.54 & 0.00 \\
            Useful-OEQ & 1.01 & 0.90 & -0.74 & -0.19 \\\cmidrule{1-5}
            Average & 0.49 & 0.52 & -0.17 & 0.01 \\
            Nature & 1.49 & 1.06 & -3.14 & -0.31 \\
            \bottomrule
        \end{tabular}
        \caption{Fine tuning effect size on each aspect, measured with BLV assessment.}
        \label{tab:blv_tn}
    \end{minipage}%
    \hfill
    \begin{minipage}[t]{0.29\textwidth}
        \centering
        \tiny
        \begin{tabular}{l>{\columncolor{gray!15}}rrr}
            \toprule
            & \multicolumn{3}{c}{\textbf{Guided Generation Effect Size}} \\\cmidrule{2-4}
            \textbf{Aspect} & GPT & 2B Base & 2B DPO \\ \midrule
            Succinct & 0.18 & -0.17 & 0.17 \\
            Diverse & -0.13 & -0.13 & 0.47 \\
            Useful-Sum & 0.48 & -0.17 & 0.57 \\
            Useful-MCQ & 0.13 & -0.20 & 0.92 \\
            Useful-OEQ & 0.76 & -0.07 & 0.77 \\\cmidrule{1-4}
            Average & 0.28 & -0.15 & 0.58 \\
            Nature & 0.33 & 0.08 & 3.17 \\
            \bottomrule
        \end{tabular}
        \caption{Guided generation effect size on each aspect, measured with BLV assessment.}
        \label{tab:blv_gg}
    \end{minipage}
\end{table*}

\section{Discussion}\label{sec:discussion}

Tables~\ref{tab:blv_cr}, ~\ref{tab:blv_tn}, and ~\ref{tab:blv_gg} show Cohen's $d$, which is the size of the effect of the treatment in the respective table. Ratings on \textbf{Nature} are not included in the average computation since it is a categorical variable; \textit{i.e.}, a low \textbf{Nature} rating simply means the description was perceived to be more straight facts-oriented than commentary-oriented, and not necessarily of a lower quality.

\paragraph{Combined Effect Size}
Table~\ref{tab:blv_cr} shows the effect size of fine tuning on \textsc{Sightation} \textit{and} applying the guided generation prompt at test time. With the combined recipe applied, the 2B model achieves an average of $0.36\sigma$ units of improvement, while the 7B model, $0.58\sigma$ units.
Intriguing observations can be made on succinctness. The 2B model exhibited the smallest effect size in this aspect, whereas the 7B model achieved the highest enhancement. This suggests that the combined recipe applied on the smaller model had negligible effect in making its descriptions more succinct. In fact, the combined recipe enhanced \textbf{Nature} by a large effect ($1.08\sigma$), implying that, with smaller models, the prime importance of the combined recipe lies in shaping the descriptions to be far more interpretive. The opposite can be said of the 7B model: the combined recipe greatly ($1.69\sigma$) enhances its succinctness, whilst shaping its descriptions far less interpretive ($-2.38\sigma$) and straight facts-oriented instead. This is in line with 3 separate comments by our BLV educators (B1, B2, and B5) who have, unknowingly of each other's interview responses, stressed the importance of succinctness: ``The description must deliver all visual items in an accurate and consistent manner, with not too long a text and including the key elements.''

\paragraph{Tuning Effect Size}
Table~\ref{tab:blv_tn} shows the effect size of fine tuning on \textsc{Sightation}. For instance, with guided generation absent, the 2B model still reaps $0.87\sigma$ units of improvement in the diversity aspect of its descriptions. The improvement margin is even amplified further by applying guided generation on the tuned model, except for usefulness in solving questions.
The table shares the observation made on the succinctness-nature relationship conveyed in Table~\ref{tab:blv_cr}, albeit to a lesser extent on the 7B model with guided generation. This set, whose ratings are on the rightmost column of Table~\ref{tab:blv_tn}, showed meaningful effect size only in usefulness as a summary and nature. This implies that larger models are already somewhat capable of capitalizing on the guided generation prompt at test time and carry less reliance on the fine tuning process.

\paragraph{Guided Generation Effect Size}
Table~\ref{tab:blv_gg} shows that the guided generation yields benefits even to GPT, possibly indicative of the under-representation of BLV accessibility needs and preferences in the pre-training data. It is important to note that, for the 2B model, the best effect of guided generation is achieved only \textit{after} the model is tuned on our dataset, again highlighting the BLV alignment capabilities of our dataset, that cannot be mimicked by test-time prompt engineering alone.

\section{Conclusion}

We release \textsc{Sightation}, a suite of the datasets showcasing these key characteristics: \textit{\romannumeral 1)} produced with BLV-oriented guided generation of VLMs instead of crowdworkers, who pose annotator bias concerns and are bottlenecked by cost and fatigue, \textit{\romannumeral 2)} validated by specialized teaching professionals at schools for the blind, and \textit{\romannumeral 3)} demonstrated across a wide range of use cases, making the most of the invaluable feedback from BLV and sighted groups and inviting continued active endeavor towards accessible language and education.

\section*{Limitations}

\paragraph{Challenges in Supervision and Capturing Details in Diagram}
One challenge of our current approach is that the supervision signal predominantly relies on the QA format, leaving the exploration of alternative supervision substances relatively underdeveloped. In addition, our pipeline does not fully leverage advanced segmentation techniques, which could be crucial for accurately capturing and interpreting complex diagrammatic details. These constraints may affect the system’s performance with diagrams that feature intricate or non-standard layouts. This aspect will be revisited in future research, as it holds the potential to achieve further advancements beyond the performance improvements demonstrated with our current dataset version.



\section*{Ethics Statement}
\paragraph{Potential Risks in Dataset Generation}
We acknowledge that during the process of creating our dataset, we utilized various LLMs, and there is a potential ethical risk that unintended biases or unexpected outcomes may have been inadvertently included. However, once the human labels are applied, the post-processed information minimizes this risk.

\paragraph{AI Assistant} Also, we hereby acknowledge that we have received assistance with grammar and word choice from LLMs such as chatGPT-4o in preparing this paper. However, all text is ultimately composed in the authors’ own words and was originally formulated by them.

\bibliography{acl_latex}

\clearpage
\onecolumn
\appendix


\section{Our Complete Dataset Collection}\label{app:rest}

We describe the rest of the dataset collection.


\subsection{\textsc{SightationVQA}}\label{app:gqa}

In constructing $\textbf{Desc}_{\texttt{++}}$ for comparison with $\textbf{Desc}_{}$, we discovered that the quality of the Question–Answer pairs directly determines the quality of the resulting context. To clarify why we invested significant effort in carefully designing these question answer pairs, we employed an LLM as a judge to evaluate and classify them according to different quality levels. To measure the quality of the Question Answer pairs, we used the VLM-as-a-Judge prompt using GPT-4o model. The prompt itself is found in Appendix~\ref{app:prompts}.


\begin{figure*}[h!]
  \centering
  \includegraphics[width=0.45\linewidth]{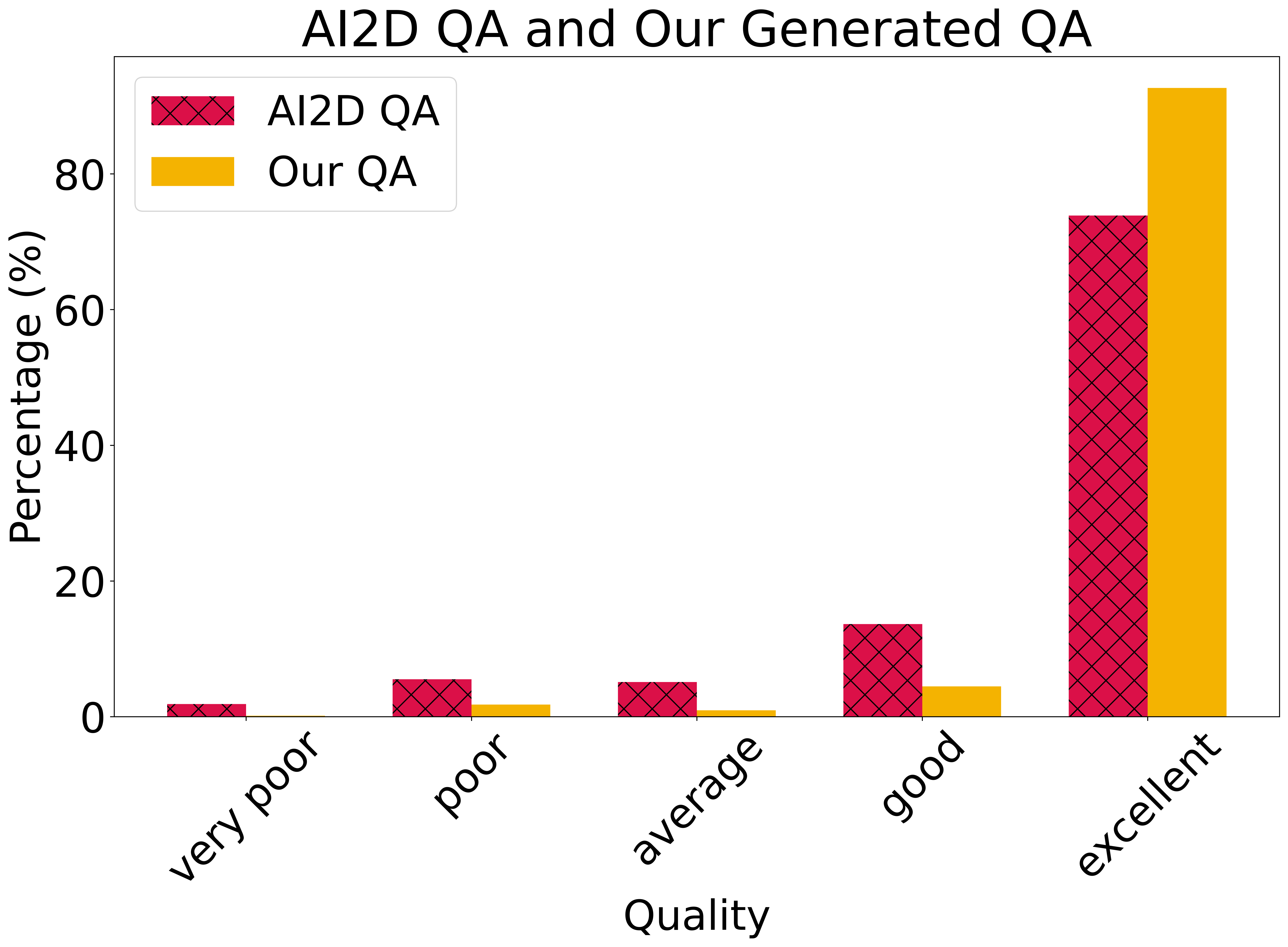}
  \caption {Percentage distribution of the quality of question-answer pairs in AI2D and \textsc{SightationVQA}}
  \label{fig:evaluate_generated_qa}
\end{figure*}

Following \citet{chen2023alpagasus} and \citet{xu2024magpie}, we compared two sets of QA pairs with \textsc{GPT-4o}. Our generated QA sets are with up to six QA pairs for each of 4,903 diagrams, producing a total of 29,438 QA pairs (sometimes exceeding six pairs per diagram). As can be seen Figure \ref{fig:evaluate_generated_qa},  we found that 92.66\% of these our generated QA pairs were rated ``Excellent'', while 4.47\% were deemed ``Good'', underscoring their high quality. By contrast, the QA pairs sourced from the AI2D dataset, though numerous, included a large portion of masked or minimally informative queries. After filtering out these masked questions, we were left with 9,708 self-contained questions spanning 3,099 diagrams, where 73.86\% received an ``Excellent'' rating and 13.65\% were deemed ``Good''. This comparison reveals that our generated QA pairs provide a more robust and contextually relevant foundation, reinforcing the value of our meticulous QA design in constructing effective $\textbf{Desc}_{\texttt{++}}$.

\subsection{\textsc{SightationReasoning}}
Employing $\textbf{Desc}$ and $\textbf{Desc}_{\texttt{++}}$, we constructed \textsc{SightationReasoning}, a reasoning dataset that consists of reasoning path and reasoning QA pairs. The prompts used for the construction of reasoning datasets are found in Appendix~\ref{app:prompts}. To verify the quality of contents as a reasoning dataset, 10\% of the samples were randomly selected to be manually inspected.

\paragraph{Reasoning Path} The reasoning path explains the logical flow or deployment of the contents in a diagram such as cause-effect relationships, step-by-step processes, explanations of phenomena, comparions of contrasts, or dependencies between components. Employing 1k diagram images and descriptions in \textsc{Sightation}, the reasoning path was identified and generated by QVQ-72B-Preview. The reasoning path extracted from $\textbf{Desc}$ and $\textbf{Desc}_{\texttt{++}}$ is denoted as $\textbf{RPath}$ and $\textbf{RPath}_{\texttt{++}}$ respectively. Consequently, one diagram possesses two reasoning paths, resulting in 2k paths in total. 

\paragraph{Reasoning QA}
The reasoning QA encompasses five types of QA pairs that require a logical understanding of diagram contents and reasoning capabilities: Causal, Process, Conditional, Explanatory, and Reverse. Similarly to the reasoning path data, $\textbf{RQA}$ and $\textbf{RQA}_{\texttt{++}}$ were generated by \textsc{QVQ-72B-Preview} using 1k diagram images and descriptions. As a result, one diagram contains 10 reasoning QA pairs in which $\textbf{RQA}$ and $\textbf{RQA}_{\texttt{++}}$ respectively include 5 pairs. While \textsc{SightationVQA} covers the visual structure and details of a diagram, the reasoning QA in \textsc{SightationReasoning} consists of more knowledge-intensive questions that require logical thinking, paving the way for the reasoning applications of \textsc{Sightation}. 

\paragraph{Evaluation} The reasoning path of \textsc{SightationReasoning} can be used as an overall representation of "logical flow" or "relationships between instances" in a diagram when understanding it, which was emphasized in the BLV educator questionnaire. To make a model employ this information when responding to reasoning questions and evaluate the reasoning paths, we fed \textsc{Qwen2-VL-7B-Instruct} with $\textbf{RPath}$ and $\textbf{RPath}_{\texttt{++}}$ separately and asked it to solve 10 questions in $\textbf{RQA}$ and $\textbf{RQA}_{\texttt{++}}$. The similarity score between the gold answers and generated answers was calculated using BERTSCore \cite{zhang2019bertscore}, and the scores for the two cases both resulted in 0.975, verifying the equal usefulness of $\textbf{RPath}$ and $\textbf{RPath}_{\texttt{++}}$.

\section{Further Related Work}

In Table~\ref{tab:dataset_comparison_extended}, we extend Table~\ref{tab:dataset_comparison} for a more comprehensive view of neighboring datasets. To the best of our knowledge, there exists no dataset to date surpassing our contribution in terms of the breadth of use cases and granularity of validation with BLV individuals.

\begin{table*}[t]
    \centering
    \small
    \definecolor{blue}{RGB}{16,85,154}
    \definecolor{pink}{RGB}{219, 76, 119}
    \newcommand{\checkblue}{\textcolor{blue}{\checkmark}}
    \newcommand{\checkpink}{\textcolor{pink}{\texttimes}}
    
    \begin{tabularx}{\textwidth}{p{4.3cm} p{1.7cm} p{1.6cm} p{2.2cm} X}
        \toprule
        \makecell[c]{\textbf{Dataset}} & \textbf{\makecell[c]{Average \\Text \\Length}} & \textbf{\makecell[c]{Validated \\by\\ BLV?}} & \makecell[c]{\textbf{Applications}} & \textbf{\makecell[c]{Dimensions\\Assessed}}\\
        \midrule
        \makecell[l]{\textbf{\textsc{Sightation} (Ours)}\\
            \hspace{0.6cm}\textsc{-Completions}\\
            \hspace{0.6cm}\textsc{-Preference}\\
            \hspace{0.6cm}\textsc{-Retrieval}\\
            \hspace{0.6cm}\textsc{-VQA}\\
            \hspace{0.6cm}\textsc{-Reasoning}
        } & \makecell[r]{\textbf{188.3}\\(words)} & \makecell[c]{\textbf{\checkblue}} & \makecell[l]{$\cdot$ \textbf{Completion} \\ $\cdot$ \textbf{Preference}\\~~~\textbf{alignment} \\$\cdot$ \textbf{Retrieval} \\ $\cdot$ \textbf{Reward} \\~~~\textbf{modeling}} & \makecell[l]{$\cdot$ \textbf{Factuality} \\ $\cdot$ \textbf{Informativeness} \\ $\cdot$ \textbf{Succinctness} \\ $\cdot$ \textbf{Diversity} \\ $\cdot$ \textbf{Usefulness}, \\~~~in 4 finer aspects \\ $\cdot$ \textbf{Interpretiveness}} \\
        \midrule
        VisText \cite{tang2023vistext} & \makecell[r]{74.6} & \makecell[c]{\checkpink} & Completion & Accuracy, Descriptiveness \\
        MathVista \cite{lu2023mathvista} & \makecell[r]{58.0} & \makecell[c]{\checkpink} & VQA, Reasoning & Correctness \\
        ChartGemma \cite{masry2024chartgemma} & \makecell[r]{37.5} & \makecell[c]{\checkpink} & Completion & Informativeness, Factual Cor-
rectness, Structure \\
        CBD \cite{bhushan-lee-2022-block} & \makecell[r]{114.5} & \makecell[c]{\checkpink} & Summarization & Adequacy, Fluency, Coherence \\
        VizWiz-VQA \cite{gurari2018vizwiz} & \makecell[r]{8.6} &  \makecell[c]{\checkblue} & VQA & Diversity, Answerability \\
        VizWiz-LF \cite{huh2024long} & \makecell[r]{73.2} & \makecell[c]{\checkblue} & VQA & Relevance, Helpfulness, Plausi-
bility, Fluency, Correctness \\
        DiagramQG \cite{zhang2024diagramqg} & \makecell[r]{9.5} & \makecell[c]{\checkpink} & DQA & Diversity, Object Density \\
        ScienceQA \cite{lu2022learn} & \makecell[r]{119.7} & \makecell[c]{\checkpink} & VQA, Reasoning & Correctness \\
        ChartQA \cite{masry-etal-2022-chartqa} & \makecell[r]{13.0} & \makecell[c]{\checkpink} & VQA & Syntactic Diversity \\
        Flickr8K \cite{hodosh2013framing} & \makecell[r]{11.8} & \makecell[c]{\checkpink} & Description & Diversity \\
        PASCAL-50S \cite{vedantam2015cider} & \makecell[r]{8.8} & \makecell[c]{\checkpink} & Description & Factuality, Literality, Generality \\
        Polaris \cite{wada2024polos} & \makecell[r]{11.5} & \makecell[c]\checkpink & Description & Fluency, Relevance, Descriptiveness \\
        Multimodal Arxiv \cite{li2024multimodal} & \makecell[r]{49.7} & \makecell[c]{\checkpink} & Description, VQA, Reasoning & Factual Alignment, Visual Clarity, Unambiguous Textual Information, Question and Option Relevance, Comprehensive Integration, Equitable Content \\
        MMMU \cite{Yue_2024_CVPR} & \makecell[r]{53.2} & \makecell[c]{\checkpink} & VQA, Reasoning & Difficulty, Knowledge, Reasoning \\
        \bottomrule
    \end{tabularx}
    \caption{Extended related work.}
    \label{tab:dataset_comparison_extended}
\end{table*}

\begin{figure}[t]
    \centering
    \includegraphics[width=0.45\linewidth]{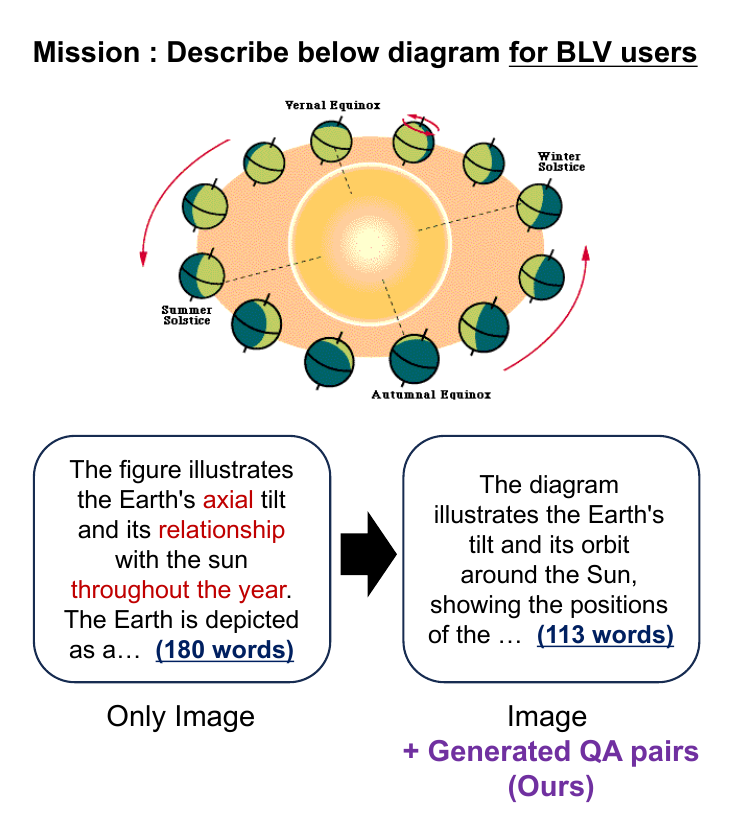}
    \caption{Less can be more for BLV users. Our approach streamlines details to highlight the core information while emphasizing key details to increase information density and maximize information efficiency per unit length.}
    \label{fig:abstract_img}
\end{figure}

\begin{table*}[htbp!]
\centering

\begin{tabular}{ll}
\toprule
Notation & Description \\ \cmidrule{1-2}
$(\cdot)^{model}$      & \pbox{0.7\textwidth}{The description \textbf{Desc} generated by (or an annotation on a generation from) a $model \in \{\texttt{g}, \texttt{q}\}$, for \textsc{GPT-4o mini} and \textsc{Qwen2-VL}, respectively. Later overloaded with narrower descriptors, such as \texttt{base}, \texttt{sft}, and \texttt{sft+dpo} to refer to the baseline/tuned models.} \\ \cmidrule{1-2}

$(\cdot)_{anchor}$      & \pbox{0.7\textwidth}{The conditioning input at the description generation stage. $anchor \in \{\texttt{None}, \texttt{++}\}$, for the one-pass image-only conditioning and the two-pass image+QA conditioning, respectively.}  \\ \cmidrule{1-2}

$\textbf{Preference}^{model}$  & \pbox{0.7\textwidth}{Preference annotation between two $\textbf{Desc}^{model}$'s on different conditioning inputs. Value takes either of the $anchor$ set \{\texttt{None}, \texttt{++}\}}  \\ \cmidrule{1-2}

$Aspect^{model}_{anchor}$      & \pbox{0.7\textwidth}{Rating annotation in terms of $Aspect \in$ \{\textbf{Factuality}, \textbf{Informativeness}, \textbf{Succinctness}, \textbf{Diversity}, \textbf{Usefulness-Gen}, \textbf{Usefulness-Sum}, \textbf{Usefulness-MCQ}, \textbf{Usefulness-OEQ}, \textbf{Nature}\}, for a description generated by $model$ conditioned on $anchor$. Value is an integer ranging from 1 to 5, on the 5-point Likert scale.}   \\ \cmidrule{1-2}

$\textbf{Best}^{model}_{anchor}$ & \pbox{0.7\textwidth}{Best sentence annotation. Value is a substring of $\textbf{Desc}^{model}_{anchor}$.}  \\

\bottomrule
\end{tabular}
\caption{Notations}
\label{tab:notation}
\end{table*}

\section{Details on the Annotations}\label{app:ann_detail}

\subsection{Logistics}

All experimentation was reviewed and approved by the Institutional Review Board. Recruiting the sighted general group was done via an online forum. Each sighted general group annotator was paid an approximate equivalent of USD80 for completing the assigned task. Recruiting the educators was done by directly corresponding with the schools for the blind. A sighted educator was compensated an approximate equivalent of USD80. A BLV educator was compensated an approximate equivalent of USD80 to USD160, depending on the number of samples completed.

\subsection{Annotations Statistics}
\paragraph{Preliminaries}

Of the 1,000 diagrams distributed to the annotators, 956 have been annotated by three annotators; 41 by two; 1 by a single annotator; and 2 by none. We collected annotations on 3,992 diagram-description pairs, each with at most 3 annotations.


\paragraph{Internal Consistency}

In Table~\ref{tab:cronbach}, we report the Cronbach's alpha value for each assessment group. The statistic is widely interpreted as the reliability of a set of survey items.

\begin{table*}[h!]
\centering

\begin{tabular}{lr}
\toprule
\textbf{Group}                   & \textbf{Cronbach's $\alpha$} \\ \cmidrule{1-2}
Sighted General         & 0.70 \\
Sighted Educators       & 0.94 \\
BLV Educators           & 0.80 \\

\bottomrule
\end{tabular}
\caption{Our survey items are considered of acceptable ($\ge0.7$) to excellent ($\ge0.9$) reliability.}
\label{tab:cronbach}
\end{table*}

\paragraph{Point-Biserial Correlation}

We examine the relationship between the binary variable, \textbf{Preference}, and the 5-point scale ratings per aspect.

\begin{table*}[h!]
\centering

\begin{tabular}{lrrrrr}
\toprule
                        & \multicolumn{5}{c}{\textbf{Aspects}} \\ \cmidrule{2-6}
\textbf{Group}          & \textbf{Factuality} & \textbf{Informativeness} & \textbf{Succinctness} & \textbf{Diversity} & \textbf{Usefulness-Gen} \\ \cmidrule{1-6}
Sighted General         & $0.36^{***}$ & $0.37^{***}$ & $0.31^{***}$ & $0.34^{***}$ & $0.43^{***}$ \\
Sighted Educators       & $0.25^{***}$ & $0.30^{***}$ & $0.30^{***}$ & $0.34^{***}$ & --- \\
\bottomrule
\end{tabular}
\caption{Correlation values between preference choice and aspect ratings were found to be moderately positive and statistically significant. (***: $p < 0.001$)}
\label{tab:point_biserial}
\end{table*}

\paragraph{Cohen's $d$}

Cohen's $d$ is a widely used statistic to measure the size of the effect of a treatment. It is the difference in the means of the treatment and control groups, normalized by the pooled standard deviation. By guidelines set forth by Cohen himself, values over 0.2 are typically considered a small effect size; 0.5, medium; and 0.8, large.

\begin{figure*}[htbp!]
    \centering
    \includegraphics[width=0.45\textwidth]{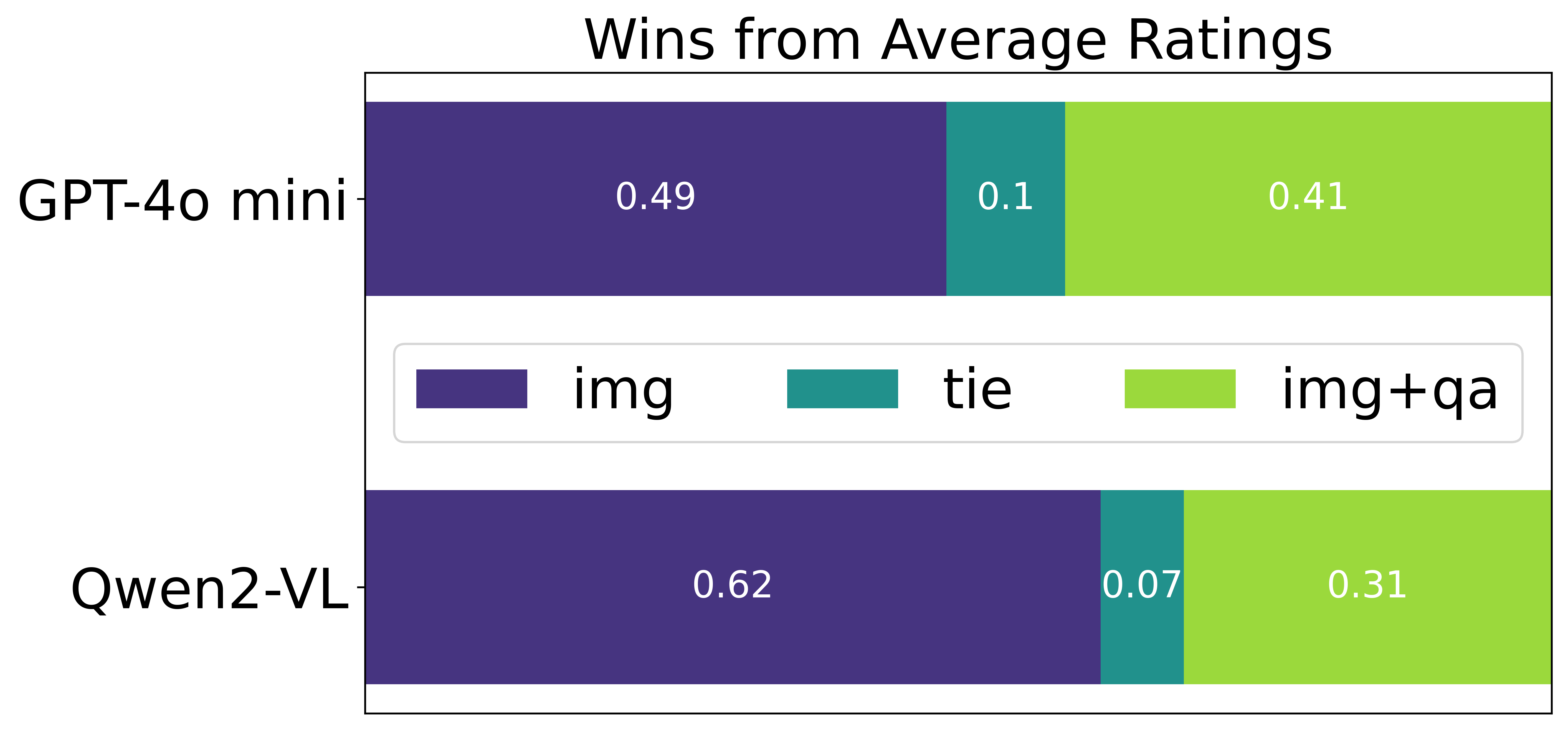}
    \caption{Win rates by $model$.}
    \label{fig:overall_wins}
\end{figure*}

\subsection{Annotations Post-processing}

\paragraph{Preference Choice}
We aggregate the multiple annotations on the basis of majority. That is, for the three-annotation samples, a 3:0 or 2:1 is considered a ``victory'' and the victor \textbf{Desc} wins that sample. For two-annotation samples with differing preferences, a tie is recorded. The overall win-loss statistics normalized against the number of diagrams (998) is shown in Figure~\ref{fig:overall_wins}.

\paragraph{Rating Assessment}


\subparagraph{Best Sentence Choice}

The best sentence for each context was manually selected by BLV annotators after listening to the context. We analyzed people’s preferences by examining the position and length of the best sentence within each context.

\subparagraph{Position}

\begin{figure*}[htbp!]
  \
  \includegraphics[width=0.48\linewidth]{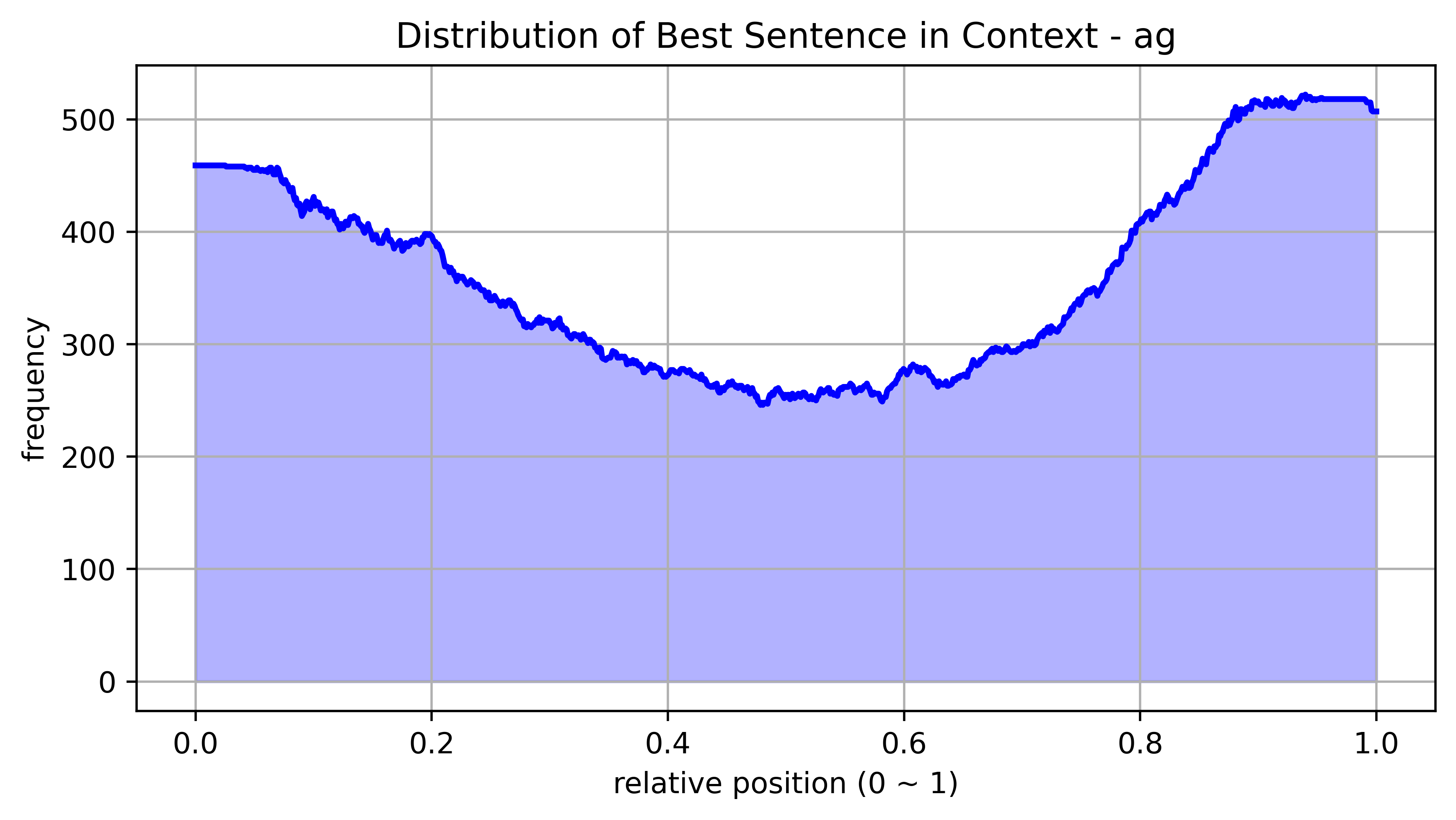} \hfill
  \includegraphics[width=0.48\linewidth]{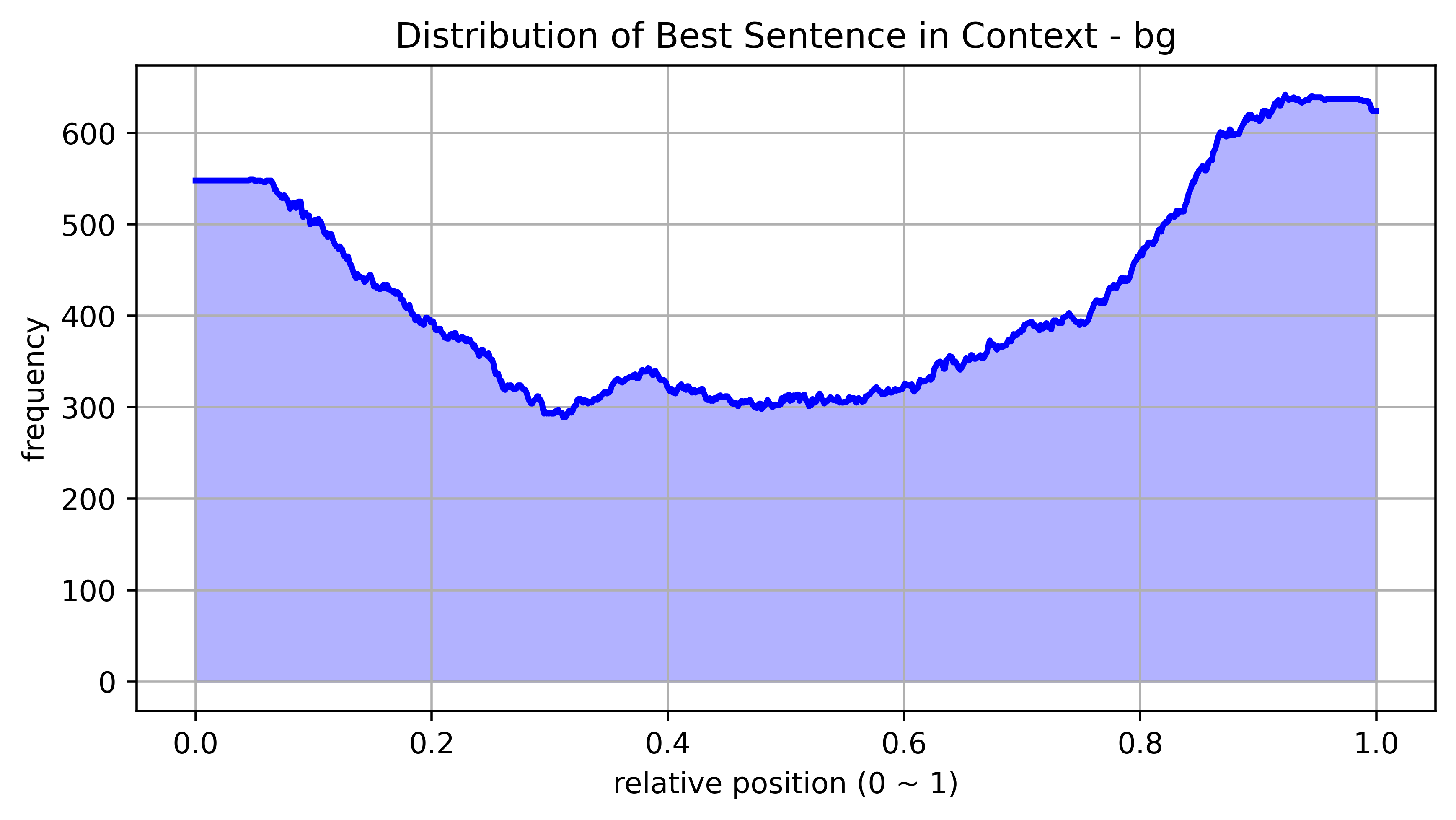}
  \caption {Descriptions generated by \textsc{GPT-4o mini}}
  \label{fig:best_sentence_distribution_g}
\end{figure*}

\begin{figure*}[htbp!]
  \
  \includegraphics[width=0.48\linewidth]{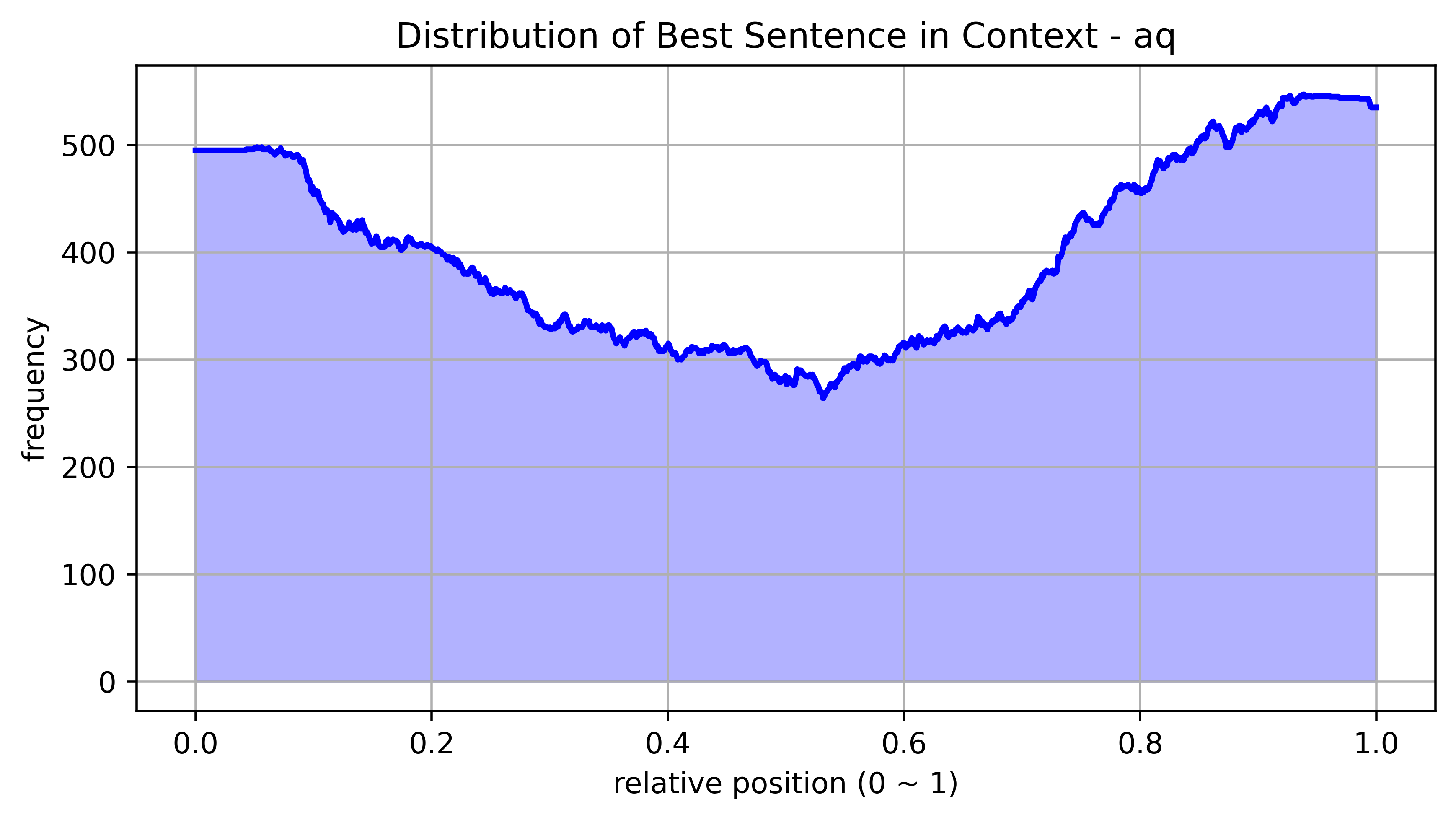} \hfill
  \includegraphics[width=0.48\linewidth]{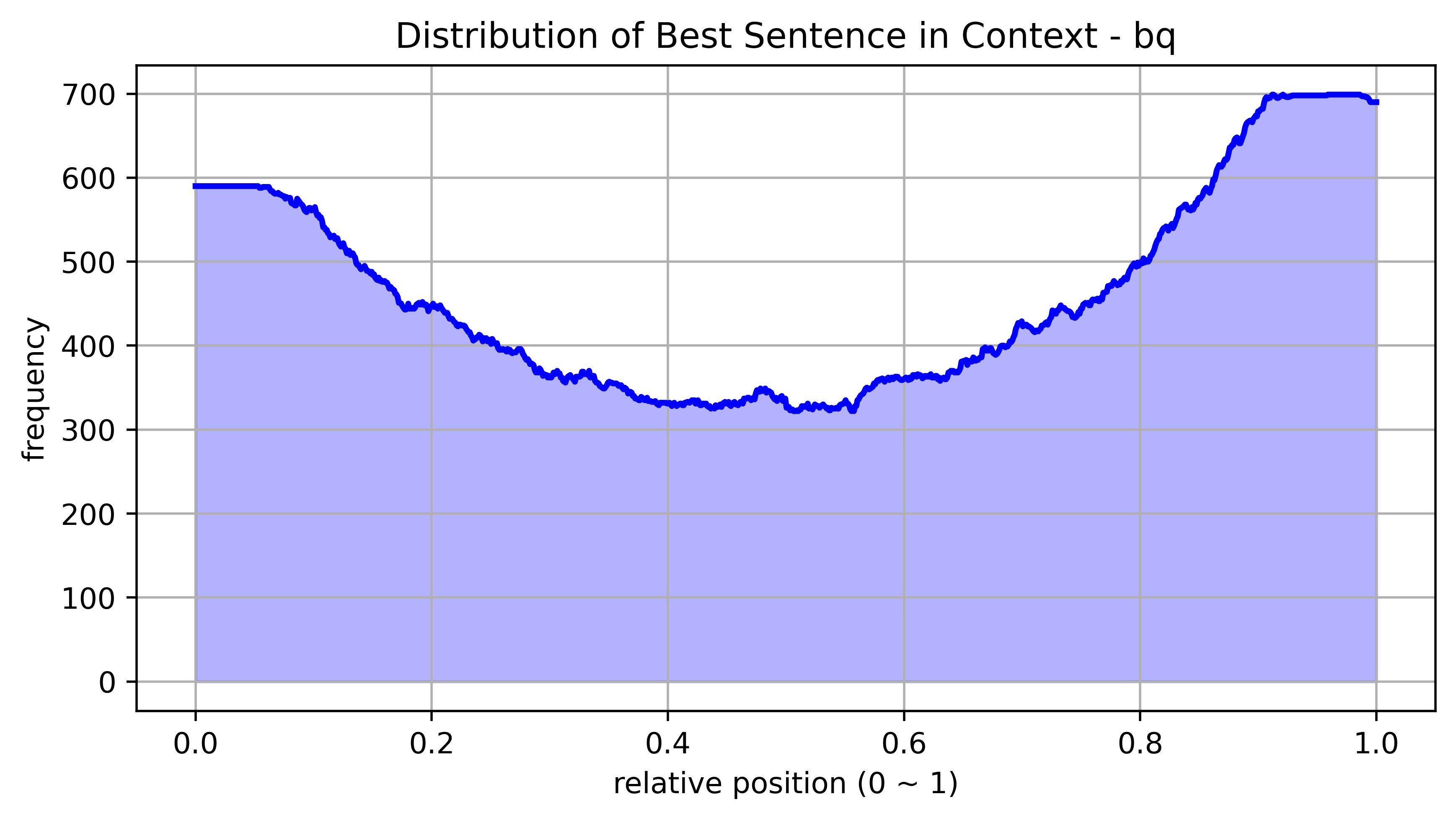}
  \caption {Descriptions generated by \textsc{Qwen2-VL}}
  \label{fig:best_sentence_distribution_q}
\end{figure*}

The normalized position of the best sentence is shown in Figures~\ref{fig:best_sentence_distribution_g}-\ref{fig:best_sentence_distribution_q}. To calculate the relative position, both the context and the best sentence were tokenized at the word level, and the position of the overlapping best sentence within the context was identified. This position was then normalized to a value between 0 and 1 by dividing it by the total length of the context. Furthermore, since some BLV annotators could not select a best sentence within the context, a filtering step was applied by setting an overlap threshold of 0.9 to account for such cases.

Figures~\ref{fig:best_sentence_distribution_g}-\ref{fig:best_sentence_distribution_q} illustrate that the best sentences in each context are predominantly positioned at the beginning and end. This pattern can be attributed to cognitive biases, specifically primacy bias and recency bias. Primacy bias refers to the tendency to place greater importance on the first pieces of information encountered in a sequence, while recency bias reflects the tendency to prioritize the most recently encountered information. Consequently, these biases increase the likelihood that preferred sentences will be selected from the beginning and end of the context.

\subparagraph{Length}

\begin{figure}[htbp!]
    \centering
    \includegraphics[width=0.48\linewidth]{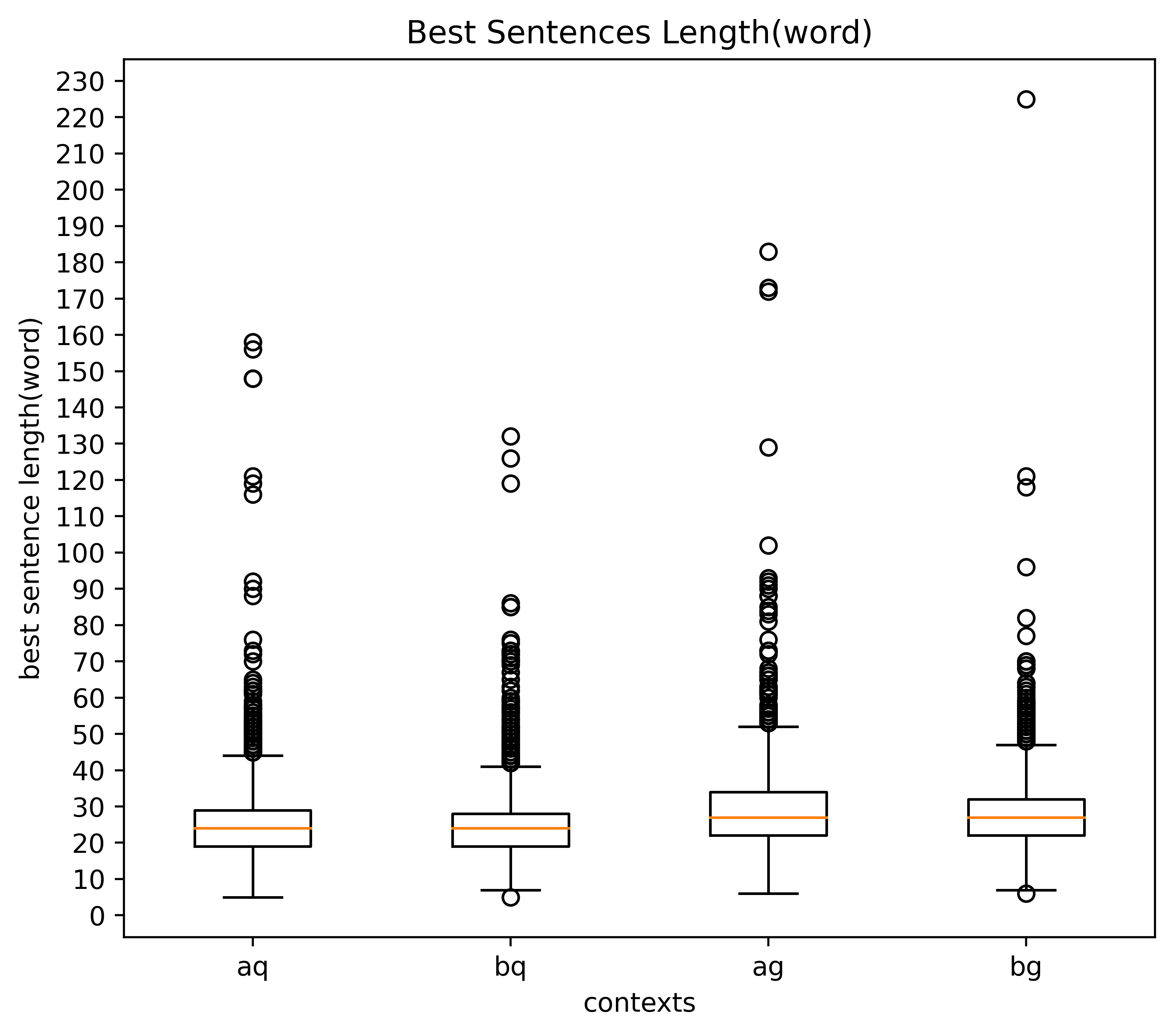}
    \caption{boxplot of best sentence length}
    \label{fig:best_sentence_length_distribution}
\end{figure}

The length of the best sentence in each context is presented in Figure~\ref{fig:best_sentence_length_distribution}. The length was determined by counting the total number of words in the best sentence. As shown in Figure 10, the best sentences across different contexts predominantly consist of 20 to 30 words, exhibiting a similar distribution pattern.

\section{Retrieval Dataset Construction}\label{app:retrieval}

The winner among the four human-annotated descriptions was assigned as the top 1 positive in terms of preference and average rating. The top 5 set contains all 4 human-annotated descriptions and 1 synthesized description; the top 10 set is a superset of the top 5, joined by 5 more synthetic descriptions. The synthetic descriptions are perturbed versions of the human-annotated descriptions, each missing a random, non-best sentence. The 10 hard negatives for an image were selected among the combined pool of top 1 descriptions for other images, sorted by cosine similarity in the embedding space. The embeddings were computed by a widely used sentence transformer, \textsc{all-mpnet-base-v2}\citep{song2020mpnet}.

\section{Detailed Results}\label{app:results}

We report the VLM-as-a-Judge evaluation and classic metric results in Tables~\ref{tab:big1a}, ~\ref{tab:big1b}, ~\ref{tab:big2B}, and ~\ref{tab:big7B}.

\subsection{Evaluation by Automatic Metrics}

\paragraph{\textsc{QVQ-72B-Preview}}

On GPT and Qwen 72B generations, the VLM judge did not reveal significant difference between the two anchors, and the little differences present aligned with assessments by the sighted general group, as can be expected from a general-purpose VLM.

It is important to note that even a state-of-the-art VLM fails to capture the BLV perspectives in text evaluation. 

\paragraph{Classic Metrics}

To our surprise, almost all instances of classic metric evaluations resulted in a win for the \texttt{++} anchor. However, the numbers from classic metrics evaluation are more of a shortcoming on the part of the classic metrics, rather than an accurate portrayal of the effectiveness of our proposed latent supervision. This is because our ``gold'' ground truths from BLV educators show that, while the QA-guided generation does manifest in ways beneficial to BLV individuals, classic automatic metrics poorly represent the assessment space covered by BLV, such as with the \textbf{Diversity} and \textbf{Usefulness-OEQ} aspects.

\begin{table*}
    \centering
    \begin{tabular}{llll}
    \toprule
    Experiment ID & & \multicolumn{2}{c}{\textbf{Assessments for}} \\ \cmidrule{3-4}
    \textbf{Description Generators}         & \textbf{Metrics} & $\textbf{Desc}_{}$ & $\textbf{Desc}_{\texttt{++}}$ \\ \midrule
    \multirow{28}{0.2\textwidth}{Experiment 1a\\
    \textsc{GPT-4o mini}\\vs.\\\textsc{GPT-4o mini}}
    
        & CLIP Score                                    & 0.476 & \textbf{0.524} \\ 
        & SigLIP Score                                  & \textbf{0.921} & 0.914 \\ 
        & BLIP-2 Retrieval Score                        & 0.495 & \textbf{0.505} \\ 
        & Self-BLEU                                     & 0.256 & \textbf{0.268} \\ 
        & PAC-Score                                     & 0.699 & \textbf{0.703} \\ 
        & LongCLIP-B Score                                & \textbf{0.507} & 0.493 \\
        & LongCLIP-L Score                                & \textbf{0.531} & 0.469 \\\cmidrule{2-4}                                    

        & $\cdot$ VLM-as-a-Judge Evaluation Average     & \textbf{4.080} & 4.033 \\
        & Factuality                                    & 4.433 & \textbf{4.445} \\
        & Informativeness                               & \textbf{4.200} & 4.166 \\
        & Succinctness                                  & 4.108 & \textbf{4.146} \\
        & Diversity                                     & \textbf{3.578} & 3.375 \\\cmidrule{2-4}

        & $\cdot$ Sighted General Group Average         & \textbf{3.983} & 3.962 \\
        & Factuality                                    & \textbf{4.128} & 4.093 \\
        & Informativeness                               & \textbf{4.367} & 4.032 \\
        & Succinctness                                  & 3.556 & \textbf{4.040} \\
        & Diversity                                     & \textbf{3.879} & 3.685 \\

        & $\cdot$ Sighted Educator Group Average        & 3.22 & \textbf{3.35} \\
        & Factuality                                    & \textbf{3.35} & 3.30 \\
        & Informativeness                               & 3.43 & 3.43 \\
        & Succinctness                                  & 2.78 & \textbf{3.53} \\
        & Diversity                                     & \textbf{3.18} & 3.08 \\
        & Usefulness to BLV                             & 3.35 & \textbf{3.40} \\
        
        & $\cdot$ BLV Educator Group Average            & 2.98 & \textbf{3.17} \\
        & Succinctness                                  & 2.43 & \textbf{2.55} \\
        & Diversity                                     & \textbf{3.23} & 3.15 \\
        & Usefulness, Summary                           & 2.95 & \textbf{3.33} \\
        & Usefulness, Multiple-chioce Questions         & 3.20 & \textbf{3.28} \\
        & Usefulness, Open-ended Questions              & 2.88 & \textbf{3.13} \\
        & Nature of Context                             & 2.98 & 3.17 \\

    \bottomrule
    \end{tabular}
\caption{The full evaluation on descriptions by GPT. Nature of Context values are not in bold because it is a categorical variable.}\label{tab:big1a}
\end{table*}

\begin{table*}
    \centering
    \begin{tabular}{llll}
    \toprule
    Experiment ID & & \multicolumn{2}{c}{\textbf{Assessments for}} \\ \cmidrule{3-4}
    \textbf{Description Generators}         & \textbf{Metrics} & $\textbf{Desc}_{}$ & $\textbf{Desc}_{\texttt{++}}$ \\ \midrule
        
    \multirow{28}{0.2\textwidth}{Experiment 1b\\
    \textsc{Qwen2-VL-72B-Instruct}\\vs.\\\textsc{Qwen2-VL-72B-Instruct}}
    
        & CLIP Score                                    & 0.451 & \textbf{0.549} \\ 
        & SigLIP Score                                  & 0.911 & \textbf{0.932} \\ 
        & BLIP-2 Retrieval Score                        & 0.494 & \textbf{0.506} \\ 
        & Self-BLEU                                     & 0.260 & \textbf{0.274} \\ 
        & PAC-Score                                     & 0.709 & \textbf{0.716} \\
        & LongCLIP-B                                    & 0.443 & \textbf{0.610} \\
        & LongCLIP-L                                    & 0.468 & \textbf{0.532} \\\cmidrule{2-4}

        & $\cdot$ VLM-as-a-Judge Evaluation Average     & \textbf{4.094} & 3.916 \\
        & Factuality                                    & \textbf{4.483} & 4.428 \\
        & Informativeness                               & \textbf{4.239} & 3.952 \\
        & Succinctness                                  & 4.026 & \textbf{4.072} \\
        & Diversity                                     & \textbf{3.629} & 3.210 \\ \cmidrule{2-4}

        & $\cdot$ Sighted General Group Average         & \textbf{4.002} & 3.850 \\
        & Factuality                                    & 3.982 & \textbf{4.060} \\
        & Informativeness                               & \textbf{4.233} & 3.782 \\
        & Succinctness                                  & 3.889 & \textbf{4.035} \\
        & Diversity                                     & \textbf{3.905} & 3.523 \\

        & $\cdot$ Sighted Educator Group Average        & 4.01 & \textbf{4.13} \\
        & Factuality                                    & 4.05 & 4.05 \\
        & Informativeness                               & \textbf{4.38} & 4.13 \\
        & Succinctness                                  & 3.80 & \textbf{4.48} \\
        & Diversity                                     & 3.80 & \textbf{3.83} \\
        & Usefulness to BLV                             & 4.03 & \textbf{4.15} \\


    \bottomrule
    \end{tabular}
\caption{The full evaluation on descriptions by the 72B model. Due to limited recruiting, BLV annotators were not given this set.}\label{tab:big1b}
\end{table*}


\begin{landscape}
\begin{table}
    \small
    \centering
    \captionsetup{width=\textwidth}
    \begin{tabular}{l>{\columncolor{gray!15}}l>{\columncolor{gray!15}}lll>{\columncolor{gray!15}}l>{\columncolor{gray!15}}l}
    \toprule
    Fine-tuning \textcolor{ColorBlue}{\textsc{Qwen2-VL-2B-Instruct}}& \multicolumn{6}{c}{\textbf{Pairwise Assessments for}  $\textbf{Desc}^{\textcolor{ColorBlue}{\texttt{q2b}}}_{}$ vs. $\textbf{Desc}^{\textcolor{ColorBlue}{\texttt{q2b}}}_{\texttt{++}}$ } \\ \cmidrule{2-7}

    \textbf{Metrics (Scores) by} & $\textbf{Desc}^{\texttt{base}}_{}$ & $\textbf{Desc}^{\texttt{base}}_{\texttt{++}}$ & $\textbf{Desc}^{\texttt{sft}}_{}$ & $\textbf{Desc}^{\texttt{sft}}_{\texttt{++}}$ & $\textbf{Desc}^{\texttt{sft+dpo}}_{}$  & $\textbf{Desc}^{\texttt{sft+dpo}}_{\texttt{++}}$ \\ \midrule

        CLIP Score                              & 0.442 & \textbf{0.558} & 0.466 & \textbf{0.534} & 0.451 & \textbf{0.549} \\ 
        SigLIP Score                            & 0.916 & \textbf{0.941} & 0.911 & \textbf{0.931} & 0.914 & \textbf{0.940} \\ 
        BLIP-2 Retrieval Score                  & 0.491 & \textbf{0.509} & 0.493 & \textbf{0.507} & 0.491 & \textbf{0.509} \\ 
        Self-BLEU                               & 0.274 & \textbf{0.278} & 0.285 & \textbf{0.291} & 0.277 & \textbf{0.281} \\ 
        PAC-Score                               & 0.711 & \textbf{0.718} & 0.706 & \textbf{0.710} & 0.712 & \textbf{0.718} \\
        LongCLIP-B                              & 0.419 & \textbf{0.581} & 0.452 & \textbf{0.548} & 0.445 & \textbf{0.555} \\ 
        LongCLIP-L                              & 0.417 & \textbf{0.583} & 0.454 & \textbf{0.546} & 0.459 & \textbf{0.541} \\ \cmidrule{1-7}     

        $\cdot$ VLM-as-a-Judge Evaluation Average & 3.307 & \textbf{3.509} & \textbf{3.732} & 3.663 & 3.334 & \textbf{3.519} \\
        Factuality                              & 3.426 & \textbf{3.783} & 3.926 & \textbf{3.974} & 3.431 & \textbf{3.784} \\
        Informativeness                         & 3.394 & \textbf{3.567} & \textbf{3.854} & 3.715 & 3.438 & \textbf{3.577} \\
        Succinctness                            & 3.346 & \textbf{3.662} & 3.707 & \textbf{3.774} & 3.347 & \textbf{3.659} \\
        Diversity                               & \textbf{3.062} & 3.025 & \textbf{3.442} & 3.188 & \textbf{3.118} & 3.054 \\ \cmidrule{1-7}

        $\cdot$ Sighted Educators Group Average & 3.91 & \textbf{3.95} & & & 4.34 & \textbf{4.49} \\
        Factuality                              & 3.95 & \textbf{4.03} & & & 4.42 & \textbf{4.66} \\
        Informativeness                         & 4.03 & \textbf{4.05} & & & 4.39 & \textbf{4.50} \\
        Succinctness                            & \textbf{3.98} & 3.90 & & & 4.37 & \textbf{4.50} \\
        Diversity                               & 3.65 & \textbf{3.80} & & & 4.18 & \textbf{4.32} \\
        Usefulness to BLV                       & 3.93 & \textbf{3.98} & & & 4.34 & \textbf{4.50} \\

        $\cdot$ BLV Educators Group Average     & \textbf{3.33} & 3.25 & \multicolumn{2}{c}{---} & 2.62 & \textbf{3.17} \\
        Succinctness                            & \textbf{3.45} & 3.33 & & & 3.15 & \textbf{3.30} \\
        Diversity                               & \textbf{3.18} & 3.10 & & & 2.03 & \textbf{2.53} \\
        Usefulness, Summary                     & \textbf{3.53} & 3.40 & & & 2.88 & \textbf{3.45} \\
        Usefulness, Multiple-choice Questions   & \textbf{3.15} & 3.10 & & & 2.88 & \textbf{3.73} \\
        Usefulness, Open-ended Questions        & 3.15 & \textbf{3.21} & & & 2.28 & \textbf{3.00} \\
        Nature of Context                       & 3.33 & 3.25 & & & 2.50 & 3.00 \\
                                    
    \bottomrule
    \end{tabular}
\caption{Evaluation of the 2B model from baseline to SFT to DPO. Note that human evaluation results are unnormalized values on the 5-point Likert scale, so direct comparisons are meaningful only within the pairwise shaded columns. SFT versus SFT samples were not distributed due to limited annotator resources. Nature of Context values are not in bold because it is a categorical variable.}\label{tab:big2B}
\end{table}
\end{landscape}

\begin{landscape}
\begin{table}
    \small
    \centering
    \captionsetup{width=\textwidth}
    \begin{tabular}{l>{\columncolor{gray!15}}l>{\columncolor{gray!15}}lll>{\columncolor{gray!15}}l>{\columncolor{gray!15}}l}
    \toprule
    Fine-tuning \textcolor{ColorOrange}{\textsc{Qwen2-VL-7B-Instruct}}& \multicolumn{6}{c}{\textbf{Pairwise Assessments for}  $\textbf{Desc}^{\textcolor{ColorOrange}{\texttt{q7b}}}_{}$ vs. $\textbf{Desc}^{\textcolor{ColorOrange}{\texttt{q7b}}}_{\texttt{++}}$ } \\ \cmidrule{2-7}

    \textbf{Metrics (Scores) by} & $\textbf{Desc}^{\texttt{base}}_{}$ & $\textbf{Desc}^{\texttt{base}}_{\texttt{++}}$ & $\textbf{Desc}^{\texttt{sft}}_{}$ & $\textbf{Desc}^{\texttt{sft}}_{\texttt{++}}$ & $\textbf{Desc}^{\texttt{sft+dpo}}_{}$  & $\textbf{Desc}^{\texttt{sft+dpo}}_{\texttt{++}}$ \\ \midrule

        CLIP Score                              & 0.423 & \textbf{0.577} & 0.411 & \textbf{0.589} & 0.407 & \textbf{0.593} \\ 
        SigLIP Score                            & 0.922 & \textbf{0.952} & 0.918 & \textbf{0.944} & 0.923 & \textbf{0.952} \\ 
        BLIP-2 Retrieval Score                  & 0.490 & \textbf{0.510} & 0.489 & \textbf{0.511} & 0.490 & \textbf{0.510} \\ 
        Self-BLEU                               & 0.268 & \textbf{0.274} & 0.275 & \textbf{0.282} & 0.268 & \textbf{0.275} \\ 
        PAC-Score                               & 0.713 & \textbf{0.720} & 0.706 & \textbf{0.714} & 0.711 & \textbf{0.718} \\
        LongCLIP-B                              & 0.419 & \textbf{0.581} & 0.452 & \textbf{0.589} & 0.417 & \textbf{0.583} \\
        LongCLIP-L                              & 0.417 & \textbf{0.583} & 0.486 & \textbf{0.514} & 0.412 & \textbf{0.588} \\ \cmidrule{1-7}  

        $\cdot$ VLM-as-a-Judge Evaluation Average & \textbf{3.951} & 3.652 & \textbf{4.021} & 3.758 & \textbf{3.948} & 3.642\\
        Factuality                              & \textbf{4.271} & 4.157 & \textbf{4.371} & 4.261 & \textbf{4.289} & 4.161 \\
        Informativeness                         & \textbf{4.101} & 3.645 & \textbf{4.161} & 3.770 & \textbf{4.100} & 3.642 \\
        Succinctness                            & \textbf{3.946} & 3.892 & \textbf{3.974} & 3.964 & \textbf{3.904} & 3.858 \\
        Diversity                               & \textbf{3.486} & 2.913 & \textbf{3.576} & 3.036 & \textbf{3.498} & 2.906 \\ \cmidrule{1-7}

        $\cdot$ Sighted Educators Group Average & \textbf{4.37} & 3.97 & & & \textbf{3.97} & 3.95 \\
        Factuality                              & \textbf{4.82} & 4.56 & & & \textbf{4.00} & 3.95 \\
        Informativeness                         & \textbf{4.67} & 3.87 & & & 4.08 & \textbf{4.13} \\
        Succinctness                            & 3.95 & \textbf{4.15} & & & 3.88 & \textbf{4.00} \\
        Diversity                               & \textbf{4.23} & 3.64 & & & \textbf{3.88} & 3.70 \\
        Usefulness to BLV                       & \textbf{4.37} & 3.97 & & & \textbf{4.03} & 3.95 \\

        $\cdot$ BLV Educators Group Average     & \textbf{3.87} & 3.82 & \multicolumn{2}{c}{---} & \textbf{3.82} & 3.71 \\
        Succinctness                            & 4.30 & \textbf{4.55} & & & 4.48 & \textbf{4.65} \\
        Diversity                               & 4.20 & 4.20 & & & \textbf{4.13} & 3.90 \\
        Usefulness, Summary                     & 4.15 & \textbf{4.55} & & & 4.25 & \textbf{4.35} \\
        Usefulness, Multiple-choice Questions   & \textbf{4.40} & 4.20 & & & \textbf{4.15} & 3.95 \\
        Usefulness, Open-ended Questions        & 3.80 & 3.80 & & & \textbf{3.70} & 3.58 \\
        Nature of Context                       & 2.35 & 1.60 & & & 2.23 & 1.85 \\
                                    
    \bottomrule
    \end{tabular}
\caption{Evaluation of the 7B model. Note that human evaluation results are nominal values on the 5-point Likert scale, so direct comparisons are meaningful only within the pairwise shaded columns. As with the 2B case, SFT versus SFT samples were not distributed due to limited annotator resources. Nature of Context values are not in bold because it is a categorical variable.}\label{tab:big7B}
\end{table}
\end{landscape}


\begin{table*}\label{tab:big3a}
    \centering
    \begin{tabular}{llll}
    \toprule
    Experiment ID & & \multicolumn{2}{c}{\textbf{Assessments for}} \\ \cmidrule{3-4}
    \textbf{Description Generators}         & \textbf{Metrics} & $\textbf{Desc}^{\textcolor{ColorBlue}{\texttt{q72bbase}}}_{}$ & $\textbf{Desc}^{\textcolor{ColorOrange}{\texttt{q7bdpo}}}_{\texttt{++}}$  \\ \midrule
    \multirow{20}{0.2\textwidth}{Experiment 3a\\\textsc{\textcolor{ColorBlue}{Qwen2-VL-72B-Instruct}}\\vs.\\\textsc{\textcolor{ColorOrange}{Fine-tuned Qwen2-VL-7B-Instruct}}} 
        & CLIP Score                                & 0.390 & \textbf{0.610} \\ 
        & SigLIP Score                              & 0.911 & \textbf{0.952} \\ 
        & BLIP-2 Retrieval Score                    & 0.487 & \textbf{0.513} \\ 
        & Self-BLEU                                 & 0.260 & \textbf{0.275} \\ 
        & PAC-Score                                 & 0.709 & \textbf{0.719} \\
        & LongCLIP-B Score                          & 0.388 & \textbf{0.612} \\
        & LongCLIP-L Score                          & 0.445 & \textbf{0.555} \\\cmidrule{2-4}                                  

        & $\cdot$ VLM-as-a-Judge Evaluation Average & \textbf{4.095} & 3.650 \\
        & Factuality                                & \textbf{4.477} & 4.238 \\
        & Informativeness                           & \textbf{4.262} & 3.586 \\
        & Succinctness                              & \textbf{3.990} & 3.894 \\
        & Diversity                                 & \textbf{3.652} & 2.880 \\\cmidrule{2-4}

        & $\cdot$ Sighted Educators Group Average   & \textbf{3.21} & 3.01 \\
        & Factuality                                & \textbf{3.30} & 3.28 \\
        & Informativeness                           & \textbf{3.33} & 2.95 \\
        & Succinctness                              & 2.95 & \textbf{3.18} \\
        & Diversity                                 & \textbf{3.13} & 2.68 \\
        & Usefulness to  BLV                        & \textbf{3.35} & 2.98 \\

        & $\cdot$ BLV Educators Group Average       & 3.69 & \textbf{4.33} \\
        & Succinctness                              & 3.60 & \textbf{4.55} \\
        & Diversity                                 & 3.60 & \textbf{3.90} \\
        & Usefulness, Summary                       & 3.95 & \textbf{4.30} \\
        & Usefulness, Multiple-choice Questions     & 3.70 & \textbf{4.55} \\
        & Usefulness, Open-ended Questions          & 3.70 & \textbf{4.45} \\
        & Nature of Context                         & 3.60 & 4.25 \\

    \bottomrule
    \end{tabular}
\caption{The smaller model outperforms a larger variant across many metrics. It is also important to note that the VLM judgments align better with sighted educators than with BLV educators. Further analysis is found in Section~\ref{sec:november}. This tendency is especially strong with the pairwise comparison between 72B- and 7B-generated descriptions. Nature of Context values are not in bold because it is a categorical variable.}
\end{table*}

\begin{table*}\label{tab:big3b}
    \centering
    \begin{tabular}{llll}
    \toprule
    Experiment ID & & \multicolumn{2}{c}{\textbf{Assessments for}} \\ \cmidrule{3-4}
    \textbf{Description Generators}         & \textbf{Metrics} & $\textbf{Desc}^{\textcolor{ColorBlue}{\texttt{q7bbase}}}_{}$ & $\textbf{Desc}^{\textcolor{ColorOrange}{\texttt{q2bdpo}}}_{\texttt{++}}$  \\ \midrule

    \multirow{20}{0.2\textwidth}{Experiment 3b \\\textsc{\textcolor{ColorBlue}{Qwen2-VL-7B-Instruct}}\\vs.\\\textsc{\textcolor{ColorOrange}{Fine-tuned Qwen2-VL-2B-Instruct}}}
    
        & CLIP Score                                & 0.486 & \textbf{0.514} \\ 
        & SigLIP Score                              & 0.922 & \textbf{0.940} \\ 
        & BLIP-2 Retrieval Score                    & 0.500 & 0.500 \\ 
        & Self-BLEU                                 & 0.268 & \textbf{0.281} \\ 
        & PAC-Score                                 & 0.713 & \textbf{0.718} \\
        & LongCLIP-B Score                          & 0.316 & \textbf{0.684} \\
        & LongCLIP-L Score                          & \textbf{0.559} & 0.441 \\\cmidrule{2-4}                                       

        & $\cdot$ VLM-as-a-Judge Evaluation Average & \textbf{3.921} & 3.545 \\
        & Factuality                                & \textbf{4.203} & 3.935 \\
        & Informativeness                           & \textbf{4.046} & 3.592 \\
        & Succinctness                              & \textbf{3.942} & 3.709 \\
        & Diversity                                 & \textbf{3.493} & 2.945 \\ \cmidrule{2-4}

        & $\cdot$ Sighted Educators Group Average   & \textbf{4.75} & 4.44 \\
        & Factuality                                & \textbf{4.75} & 4.50 \\
        & Informativeness                           & \textbf{4.65} & 4.38 \\
        & Succinctness                              & \textbf{4.88} & 4.40 \\
        & Diversity                                 & \textbf{4.80} & 4.63 \\
        & Usefulness to BLV                         & \textbf{4.65} & 4.28 \\

        & $\cdot$ BLV Educators Group Average       & 4.13 & \textbf{4.32} \\
        & Succinctness                              & 4.05 & \textbf{4.15} \\
        & Diversity                                 & 4.08 & \textbf{4.15} \\
        & Usefulness, Summary                       & 3.85 & \textbf{4.13} \\
        & Usefulness, Multiple-choice Questions     & 4.53 & \textbf{4.58} \\
        & Usefulness, Open-ended Questions          & 4.23 & \textbf{4.35} \\
        & Nature of Context                         & 4.08 & 4.50 \\

    \bottomrule
    \end{tabular}
\caption{The 2B model performs on par with the 7B variant. Again, VLM judgments align better with sighted educators than with BLV educators. Further analysis is found in Section~\ref{sec:november}. Nature of Context values are not in bold because it is a categorical variable.}
\end{table*}

\begin{table*}\label{tab:big3c}
    \centering
    \begin{tabular}{llll}
    \toprule
    Experiment ID & & \multicolumn{2}{c}{\textbf{Assessments for}} \\ \cmidrule{3-4}
    \textbf{Description Generators}         & \textbf{Metrics} & $\textbf{Desc}^{\textcolor{ColorBlue}{\texttt{chartgemma}}}_{}$ & $\textbf{Desc}^{\textcolor{ColorOrange}{\texttt{q2bsft}}}_{}$  \\ \midrule

    \multirow{10}{0.2\textwidth}{Experiment 3c \\\textsc{\textcolor{ColorBlue}{ChartGemma (3B)}}\\vs.\\\textsc{\textcolor{ColorOrange}{Fine-tuned Qwen2-VL-2B-Instruct}}}
    
        & CLIP Score                                & 0.450 & \textbf{0.550} \\ 
        & SigLIP Score                              & 0.872 & \textbf{0.940} \\ 
        & BLIP-2 Retrieval Score                    & \textbf{0.511} & 0.490 \\ 
        & Self-BLEU                                 & \textbf{0.305} & 0.280 \\ 
        & PAC-Score                                 & 0.705 & \textbf{0.716}
        \\
        & LongClip-B                                & 0.316 & \textbf{0.684}
        \\
        & LongClip-L                                & \textbf{0.559} & 0.441\\ \cmidrule{2-4}                                       

        & $\cdot$ VLM-as-a-Judge Evaluation Average & 2.951 & \textbf{3.860} \\
        & Factuality                                & 3.068 & \textbf{4.119} \\
        & Informativeness                           & 2.848 & \textbf{3.967} \\
        & Succinctness                              & 3.253 & \textbf{3.925} \\
        & Diversity                                 & 2.635 & \textbf{3.428} \\

        
    \bottomrule
    \end{tabular}
\caption{A 2B model fine-tuned on \textsc{SightationCompletions} outperforms a 3B model tuned on a larger dataset. Note that \textsc{ChartGemma} is not meant for conversational use. Hence, for a fair comparison, we did \textit{not} enter our guided generation prompt and instead input only the brief request ``Generate a caption'' to both models.}
\end{table*}

\begin{table*}[t]
    \centering
    \begin{tabular}{l>{\columncolor{gray!15}}c>{\columncolor{gray!15}}ccc>{\columncolor{gray!15}}c>{\columncolor{gray!15}}c}
    \toprule
    & \multicolumn{6}{c}{2-way Cross-validation of \textbf{BLIP-2}} \\ \cmidrule{2-7}
    \textbf{Train set} & \multicolumn{2}{c}{N/A (Pre-trained)} & \multicolumn{2}{c}{COCO} & \multicolumn{2}{c}{\textsc{SightationRetrieval} (Ours)} \\ \cmidrule(lr){2-3}\cmidrule(lr){4-5}\cmidrule(lr){6-7}
    \textbf{Test set} & COCO & Ours & COCO & Ours & COCO & Ours \\ \midrule
    Recall@1                                & 0.171 & 0.048 & 0.185 & 0.033 & 0.180 & 0.076 \\ 
    Recall@5                                & 0.767 & 0.210 & 0.831 & 0.134 & 0.766 & 0.348 \\ 
    Recall@10                               & ---   & 0.340 & ---   & 0.229 & ---   & 0.549 \\ 
    Precision@1                             & 0.856 & 0.371 & 0.924 & 0.250 & 0.900 & 0.585 \\ 
    Precision@5                             & 0.767 & 0.324 & 0.831 & 0.204 & 0.766 & 0.535 \\
    Precision@10                            & ---   & 0.263 & ---   & 0.175 & ---   & 0.425 \\ 
    \bottomrule
    \end{tabular}
    \caption{\textsc{SightationRetrieval} shows promising potential as a challenging and effective training material for image-to-text retrievers. Two important observations can be made: the model trained on our set generalizes to COCO better than the other direction; our model performs on par with the model that was both trained and tested on COCO.  $K=10$ values are missing for tests with COCO, since its samples contain only 5 positives each.}
    \label{tab:retrieval}
\end{table*}

\begin{figure}[p!]
    \centering
    \includegraphics[width=0.95\linewidth]{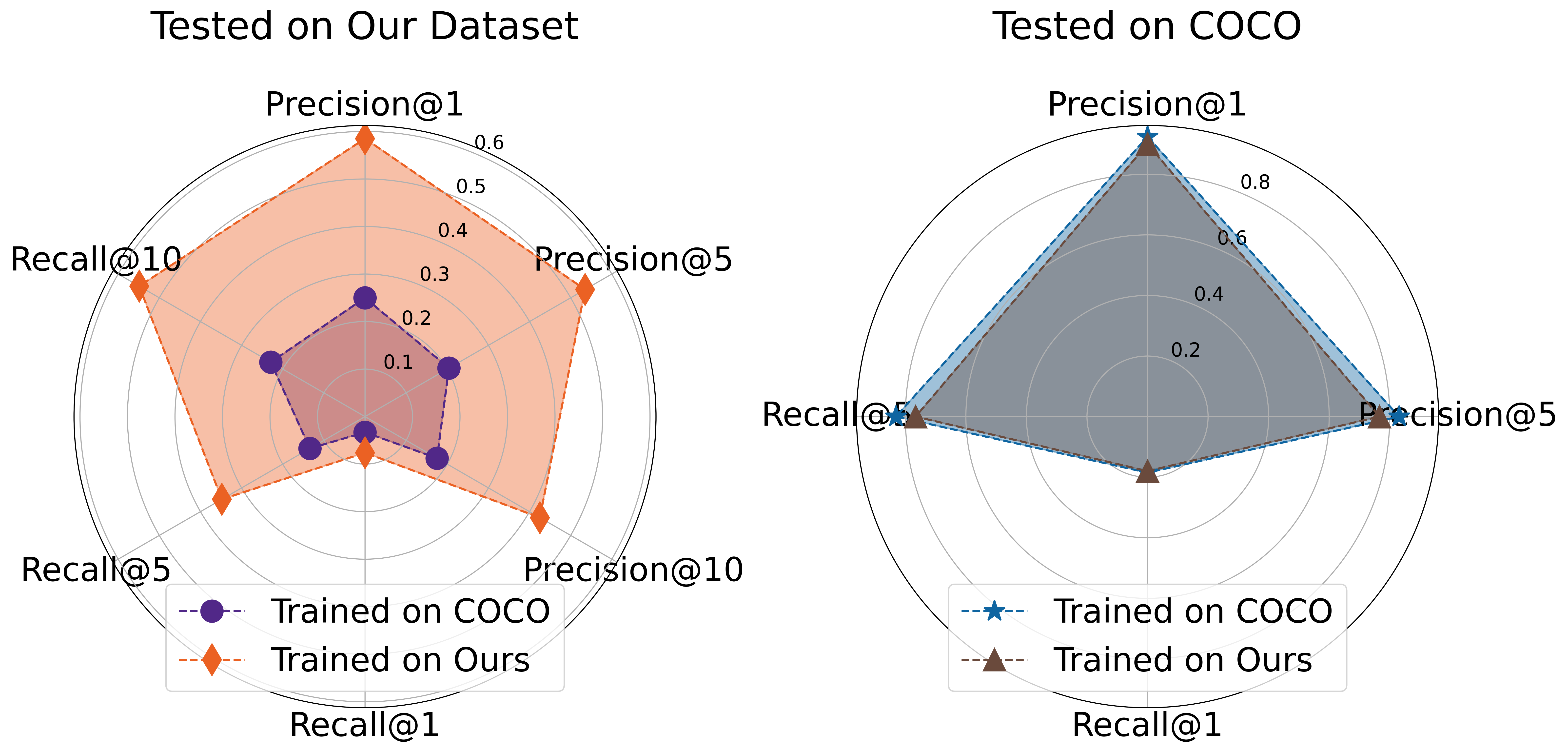}
    \caption{Retrieval performance was measured with 2-way cross validation. On our test set (Left), the COCO-tuned \textsc{BLIP-2} generalizes poorly, whereas on the COCO test set (Right), the \textsc{SightationRetrieval}-tuned \textsc{BLIP-2} performs on par with the COCO-tuned \textsc{BLIP-2}.} 
    \label{fig:retrieval_tuning}
\end{figure}

\section{Annotator Demographics and Interviews}\label{app:demo}

\subsection{Demographics}

\subsubsection{BLV Educators}
Please refer to Table~\ref{tab:blv_educators_demo}.

\begin{table}[h!]
    \centering
    \tiny
    \begin{tabular}{lccccll}
    \toprule
    \textbf{ID} & \textbf{Sex} & \textbf{Age} & \makecell[c]{\textbf{Teaching}\\\textbf{Experience}\\\textbf{(years)}} & \makecell[c]{\textbf{Onset}\\\textbf{Age}} & \makecell[c]{\textbf{AI}\\\textbf{Use,}\\\textbf{Generic}} & \makecell[c]{\textbf{AI}\\\textbf{Use,}\\\textbf{Accessibility}} \\
    \midrule
    B1 & M & 54 & 28  & 16 & ChatGPT, Gemini & SenseReader \\
    B2 & F & 46 & 21  & Congenital & ChatGPT & SenseReader \\
    B3 & M & 47 & 5  & 9 & ChatGPT, Gemini & SenseReader \\
    B4 & M & 51 & 26 & 14 & SeeingAI, ChatGPT, Adot, Perplexity, Adot & SenseReader, NVDA, VoiceOver \\
    B5 & M & 20 &  1 & Congenital & SeeingAI, ChatGPT & SenseReader, NVDA \\
    B6 & M & 46 & 19 & --- & --- & SenseReader \\
    B7 & M & 44 & 21 &  Congenital & Be\_My\_Eyes, SeeingAI, ChatGPT, Claude & SenseReader, VoiceOver \\
    B8 & M & 45 & 19 &  Congenital & Be\_My\_Eyes, SeeingAI, ChatGPT & SenseReader, VoiceOver \\
    \bottomrule
    \end{tabular}
    \caption{BLV Teachers Information. All the BLV teachers in our study were of blindness level 1, the severest.}
    \label{tab:blv_educators_demo}
\end{table}

\subsubsection{Sighted Educators}
Please refer to Table~\ref{tab:sighted_educators_demo}.

\begin{table}[h!]
    \centering
    \footnotesize
    \begin{tabular}{lcccl}
    \toprule
    \textbf{ID} & \textbf{Sex} & \textbf{Age} & \textbf{Teaching Experience (years)} & \textbf{AI Use - Generic} \\
    \midrule
    S1 & M & 39 & 6.5 & ChatGPT \\
    S2 & M & 51 & 20 & ChatGPT, wrtn \\
    S3 & M & 48 & 21 & ChatGPT \\
    S4 & F & 40 & 13 & ChatGPT \\
    S5 & F & 56 & 33 & --- \\
    S6 & F & 49 & 20 & ChatGPT \\
    S7 & M & 49 & 20 & Gemini \\
    S8 & F & 49 & 24 & ChatGPT, Claude \\
    S9 & M & 44 & 14 & --- \\
    S10 & F & 50 & 20 & ChatGPT \\
    \bottomrule
    \end{tabular}
    \caption{Sighted Teachers Information.}
    \label{tab:sighted_educators_demo}
\end{table}

\section{Prompts}\label{app:prompts}

\centering
\begin{tcolorbox}[
    colback=white,
    colframe=gray!20,
    arc=10pt,
    boxrule=1pt,
    title={\textcolor{white}{\textbf{Prompts for Context Generation}}},
    coltitle=white,
    fonttitle=\large,
    colbacktitle=black!70,
]

\paragraph{\textbf{Generating} $\textbf{Desc}_{}$}
You are a helpful expert who is knowledgeable in various fields of academia. You are skilled in reading, interpreting, and understanding academic papers and figures contained therein. You are tasked with elaborating on the given information, which consists of a figure image. Write an informative and explanatory text in one paragraph under 200 words that describes the basic characteristics of the figure and incorporates important information. You may attempt to internally identify implicit points of curiosity for someone who is trying to understand the given figure, and then include explanations for those points in your response. Avoid mere reiteration of the given information as much as possible. You need not specify the origins of various parts of your response. [Optional: Aspect Suffix]

\paragraph{\textbf{Generating} $\textbf{Desc}_{\texttt{++}}$}
You are a helpful expert who is knowledgeable in various fields of academia. You are skilled in reading, interpreting, and understanding academic papers and figures contained therein. You are tasked with elaborating on the given information, which consists of a figure image and several question-answer pairs that have been derived from the figure. Write an informative and explanatory text in one paragraph under 200 words that describes the basic characteristics of the figure and incorporates important information from the question-answer pairs. You may attempt to internally identify implicit points of curiosity for someone who is trying to understand the given figure, and then include explanations for those points in your response. Avoid mere reiteration of the given information as much as possible. You need not specify the origins of various parts of your response. Here is the reference information: [QA\_PAIRS: {vqas}] [Optional: Aspect Suffix]

\end{tcolorbox}

\centering
\begin{tcolorbox}[
    colback=white,
    colframe=gray!20,
    arc=10pt,
    boxrule=1pt,
    title={\textcolor{white}{\textbf{Aspect Suffixes}}},
    coltitle=white,
    fonttitle=\large,
    colbacktitle=black!70,
]

\paragraph{Factuality}
When generating the diagram description, pay close attention to making it factual. A highly factual text delivers only the facts that are grounded in the diagram.

\paragraph{Informativeness}
When generating the diagram description, pay close attention to making it informative. A highly informative text describes all of the diagram, holistically.

\paragraph{Succinctness}
When generating the diagram description, pay close attention to making it succinct. A highly succinct text is concise and to the point.

\paragraph{Diversity}
When generating the diagram description, pay close attention to making it diverse. A highly diverse text captures a variety of perspectives from the diagram and employs multiple effective ways of getting the diagram message across. 

\end{tcolorbox}


\centering
\begin{tcolorbox}[
    colback=white,
    colframe=gray!20,
    arc=10pt,
    boxrule=1pt,
    title={\textcolor{white}{\textbf{Prompt for Question-answer Pair Generation}}},
    coltitle=white,
    fonttitle=\large,
    colbacktitle=black!70,
]

Please generate six question-and-answer pairs based on the provided image to aid in creating a comprehensive context. This context should include all essential details, allowing BLV (Blind and Low Vision) users to rely on the generated text instead of viewing the image (\textit{e.g.}, accessing information audibly). The question-and-answer pairs should cover both the main structure and finer details present in the image.
\end{tcolorbox}

\centering
\begin{tcolorbox}[
    colback=white,
    colframe=gray!20,
    arc=10pt,
    boxrule=1pt,
    title={\textcolor{white}{\textbf{Prompts for Reasoning Path Generation}}},
    coltitle=white,
    fonttitle=\large,
    colbacktitle=black!70,
]

You will be provided with a diagram, along with two descriptions of it. As an expert and experienced educator, you are tasked to examine your descriptions to identify common reasoning paths, such as cause-effect relationships, step-by-step processes, explanations of phenomena, comparisons of contrasts, and dependencies between components.

The identified reasoning paths should be under 25 words. Please provide the reasoning paths that you examined in the following JSON format:

\begin{verbatim}
        {"Context1": {"ReasoningPath": text},
         "Context2": {"ReasoningPath": text}}
\end{verbatim}

DO NOT return anything other than the JSON above.
\end{tcolorbox}

\centering
\begin{tcolorbox}[
    colback=white,
    colframe=gray!20,
    arc=10pt,
    boxrule=1pt,
    title={\textcolor{white}{\textbf{Prompts for Reasoning QA Generation}}},
    coltitle=white,
    fonttitle=\large,
    colbacktitle=black!70,
]

You will be provided with a diagram, along with two descriptions of it. As an expert and experienced educator, you are tasked to examine your descriptions to generate reasoning question and answer pairs of five categories such as:

\begin{itemize}
    \item Causal Reasoning: "Why does [event] happen?"
    \item Process Reasoning: "What happens after [event]?"
    \item Conditional Reasoning: "What if [condition] changes?"
    \item Explanatory Reasoning: "Explain the role of [component] in the process."
    \item Reverse Reasoning: "Given [outcome], what might have caused it?"
\end{itemize}

Please provide the reasoning question and answer pairs that you generated in the following JSON format:

\begin{verbatim}
        {"Context 1": {"Causal": {"Question": text,
                                    "Answer": text},
                        "Process": {"Question": text,
                                    "Answer": text},	
                        "Conditional": {"Question": text,
                                        "Answer": text},
                        "Explanatory": {"Question": text,
                                        "Answer": text},
                        "Reverse": {"Question": text,
                                    "Answer": text}
            },
            "Context 2": {"Causal": {"Question": text,
                                    "Answer": text},
                        ..
                        "Reverse": {"Question": text,
                                    "Answer": text}
        }
\end{verbatim}

Each generated question should be under 15 words and each corresponding answer should be under 25 words.

DO NOT return anything other than the JSON above.

\end{tcolorbox}

\centering
\begin{tcolorbox}[
    colback=white,
    colframe=gray!20,
    arc=10pt,
    boxrule=1pt,
    title={\textcolor{white}{\textbf{Prompt for VLM-as-a-Judge Evaluation of Description Pairs}}},
    coltitle=white,
    fonttitle=\large,
    colbacktitle=black!70,
]
You will be provided with a diagram, along with two descriptions of it. As an expert and experienced educator, you are tasked to evaluate each description based on the following qualities on a 5-point Likert scale. For each statement, give a score corresponding to how strongly you agree with the given statement: 1 (Strongly Disagree), 2 (Disagree), 3 (Neutral), 4 (Agree), or 5 (Strongly Agree).

\begin{itemize}
    \item Diversity: The description captures a variety of perspectives from the diagram and conveys multiple effective ways of getting the diagram message across.
    \item Succinctness: The description is concise and to the point, avoiding unnecessary details.
    \item Factuality: The description is accurate and reflects solely the information presented in the diagram.
    \item Informativeness: The description covers the diagram holistically, and it effectively conveys the main trends and insights of the diagram.
\end{itemize}

Please provide your ratings in the following JSON format:

\begin{verbatim}
        {
            Context 1: {
                Diversity: score,
                Succinctness: score,
                Factuality: score,
                Informativeness: score
            },
            Context 2: {
                Diversity: score,
                Succinctness: score,
                Factuality: score,
                Informativeness: score
            }
        }
\end{verbatim}

DO NOT return anything other than the JSON above.

\end{tcolorbox}

\begin{tcolorbox}[
    colback=white,
    colframe=gray!20,
    arc=10pt,
    boxrule=1pt,
    title={\textcolor{white}{\textbf{Prompt for VLM-as-a-Judge Evaluation of Question-Answer Pairs}}},
    coltitle=white,
    fonttitle=\large,
    colbacktitle=black!70,
]
\textbf{Instruction}

You need to rate the quality of the given Question and Answer in relation to a diagram. Specifically, assess whether the Q\&A correctly references and interprets the information presented in the diagram. Consider the Q\&A’s clarity, specificity, and coherence as they pertain to the diagram’s content.You must take into account not only the Answer but also whether an appropriate Question has been provided for the given diagram at the same time.
The rating scale is as follow:

\begin{itemize}
    \item very poor: The Q\&A is unclear, vague, or incoherent in relation to the diagram. It lacks essential information or misinterprets the diagram’s content.
    \item poor: The Q\&A is somewhat unclear or omits important details from the diagram. It requires significant clarification or correction to align with the diagram.
    \item average: The Q\&A is moderately clear and specific. It may need additional details or minor clarifications to fully match the diagram’s information.
    \item good: The Q\&A is clear, specific, and mostly well-formed in referencing the diagram. It provides sufficient context to understand how the diagram supports the question and answer.
    \item excellent: The Q\&A is very clear, specific, and well-articulated. It precisely references and fully aligns with the diagram, containing all necessary details and context.
\end{itemize}

\textbf{Output Format}

Given the user’s diagram, question, and answer, you must:

Provide an assessment that briefly explains the strengths and/or weaknesses of how the Q\&A relates to the diagram.
Output your rating (one of: very poor, poor, average, good, excellent) by filling in the placeholders below.

\begin{verbatim}
[
{
	"explanation": "[...]",
	"input_quality": "[very poor/poor/average/good/excellent]"
},
...
]
\end{verbatim}

\textbf{Notes}

\begin{itemize}
\item DO NOT return anything else other than the JSON above.
\item Number of item in above list should be same as the number of given QA pairs. Also the order for the explanation and input quality should be same as input QA's order
\end{itemize}

\end{tcolorbox}

\section{Fine-tuning Configurations}\label{app:tuning_config}

\begin{table*}[t!]
\centering
\small
\begin{tabularx}{\textwidth}{lXX}
\toprule
\textbf{Parameter} & \textbf{SFT Config (Qwen2-VL-2B-Instruct)} & \textbf{DPO Config (Qwen2-VL-2B-Instruct)} \\ 
\midrule
\multicolumn{3}{l}{\textbf{Script Arguments}} \\
Dataset Name & \textsc{SightationCompletions} & \textsc{SightationPreference} \\
\midrule
\multicolumn{3}{l}{\textbf{Training Configurations}} \\
Output Directory & anonymous & anonymous \\
Evaluation Strategy & steps & steps \\
Train Batch Size & 1 & 1 \\
Evaluation Batch Size & 1 & 1 \\
Gradient Accumulation Steps & 8 & 8 \\
Training Epochs & 1 & 1 \\
Save Total Limit & 5 & 5 \\
bfloat16 Enabled & true & true \\
Evaluation Steps & 10 & 10 \\
Label Names & ["labels"] & ["labels"] \\
Load Best Model at End & true & true \\
Metric for Best Model & eval\_loss & eval\_loss \\
Use Liger & true & true \\
Max Sequence Length & 1024 & 1024 \\
Remove Unused Columns & false & true \\
Dataset Kwargs & skip\_prepare\_dataset: true & skip\_prepare\_dataset: false \\
Gradient Checkpointing & true & true \\
Gradient Checkpointing Kwargs & use\_reentrant: false & use\_reentrant: false \\
Dataset Num Processors & 8 & 8 \\
Torch Compile & true & --- \\
DDP Find Unused Parameters & --- & true \\
\midrule
\multicolumn{3}{l}{\textbf{Model Config}} \\
Use PEFT & false & false \\
Model Path & Qwen/Qwen2-VL-2B-Instruct & Qwen/Qwen2-VL-2B-Instruct \\
Torch Dtype & bfloat16 & bfloat16 \\
Attention Implementation & flash\_attention\_2 & flash\_attention\_2 \\
\bottomrule
\end{tabularx}
\caption{SFT and DPO configurations for Qwen2-VL-2B-Instruct. Tuning was performed on 4 \texttimes A6000 GPUs.
}
\label{tab:tuning_config_2b_sft_dpo}
\end{table*}

\begin{table*}[t]
\centering
\small
\begin{tabularx}{\textwidth}{lXX}
\toprule
\textbf{Parameter} & \textbf{SFT Config (Qwen2-VL-7B-Instruct)} & \textbf{DPO Config (Qwen2-VL-7B-Instruct)} \\ 
\midrule
\multicolumn{3}{l}{\textbf{Script Arguments}} \\
Dataset Name & \textsc{SightationCompletions}  &  \textsc{SightationPreference} \\
\midrule
\multicolumn{3}{l}{\textbf{Training Configurations}} \\
Output Directory & anonymous & anonymous \\
Evaluation Strategy & steps & steps \\
Train Batch Size & 1 & 1 \\
Evaluation Batch Size & 1 & 1 \\
Gradient Accumulation Steps & 8 & 8 \\
Training Epochs & 1 & 1 \\
Save Total Limit & 5 & 5 \\
bfloat16 Enabled & true & true \\
Evaluation Steps & 10 & 10 \\
Label Names & ["labels"] & ["labels"] \\
Load Best Model at End & false & false \\
Metric for Best Model & eval\_loss & eval\_loss \\
Use Liger & true & true \\
Max Sequence Length & 1024 & 1024 \\
Remove Unused Columns & false & true \\
Dataset Kwargs & skip\_prepare\_dataset: true & skip\_prepare\_dataset: false \\
Gradient Checkpointing & true & true \\
Gradient Checkpointing Kwargs & use\_reentrant: false & use\_reentrant: false \\
Dataset Num Processors & 8 & 8 \\
DDP Find Unused Parameters & true & true \\
\midrule
\multicolumn{3}{l}{\textbf{Model Config}} \\
Use PEFT & true & true \\
Model Path & Qwen/Qwen2-VL-7B-Instruct & Qwen/Qwen2-VL-7B-Instruct \\
Torch Dtype & bfloat16 & bfloat16 \\
Attention Implementation & flash\_attention\_2 & flash\_attention\_2 \\
LoRA Rank (r) & 16 & 16 \\
LoRA Alpha & 16 & 16 \\
LoRA Dropout & 0.1 & 0.1 \\
LoRA Target Modules & q\_proj, k\_proj, v\_proj, o\_proj, gate\_proj, up\_proj, down\_proj & q\_proj, k\_proj, v\_proj, o\_proj, gate\_proj, up\_proj, down\_proj \\
\bottomrule
\end{tabularx}
\caption{SFT and DPO configurations for Qwen2-VL-7B-Instruct. Tuning was performed on 4 \texttimes A6000 GPUs.
}
\label{tab:tuning_config_7b_sft_dpo}
\end{table*}

\begin{table*}[t]
\centering
\small
\begin{tabularx}{\textwidth}{lX}
\toprule
\textbf{Component} & \textbf{Configuration} \\ \midrule
\textbf{Model} & BLIP-2 (Salesforce/blip2-itm-vit-g) \\
\textbf{GPUs} & Text model on CUDA:0, Vision model on CUDA:1 \\
\textbf{Dataset} & \textsc{SightationRetrieval} \\
\textbf{Loss} & InfoNCE (temperature = 0.07) \\
\textbf{Batch Size} & 1 (with gradient accumulation steps = 4) \\
\textbf{Epochs} & 5 \\
\textbf{Optimizer} & AdamW (Text LR: 5e-5, Vision LR: 2e-5) \\
\textbf{Gradient Clipping} & Max norm = 1.0 \\
\textbf{Scheduler} & Linear warmup (10\% of steps) \\
\textbf{Frozen Layers} & All except: layernorm, projection, encoder layers 10-11 (Vision); layernorm, projection, encoder layers 10-11, crossattention (Text) \\
\textbf{Checkpoints} & Best and per-epoch saved to anonymized path \\
\bottomrule
\end{tabularx}
\caption{Training configurations for BLIP-2 image-text retrieval.}
\label{tab:retrieval_config}
\end{table*}


\clearpage
\section{Guidelines}\label{app:guidelines}

\centering
\begin{tcolorbox}[
    colback=white,
    colframe=gray!20,
    arc=10pt,
    boxrule=1pt,
    title={\textcolor{white}{\textbf{Annotation Guidelines for the Sighted General Group (1/4)}}},
    coltitle=white,
    fonttitle=\large,
    colbacktitle=black!70,
]

Please carefully read the guidelines below and ensure accurate labeling. Your responses are considered high-quality data and can have critical implications for the experiment. Pay special attention to the Caution section.

\paragraph{Annotation Guidelines}
Thank you for contributing to this project. In the following paragraphs, we will walk you through the project description, your tasks, and annotation examples.

\paragraph{Project}
Our project targets the visually impaired. People who are Blind or have Low Vision (BLV) do not always benefit from the latest AI developments in the same way or extent as sighted users. In this pilot study, we would like to first assess exactly how much state-of-the-art models may assist sighted users, so that we may gain insights into (i) what state-of-the-art models can and cannot do and (ii) what modifications might be necessary to alter their assistive information to cater specifically for BLV users.

\paragraph{Task}
Each task you are about to complete consists of:
\begin{itemize}
    \item 1 image
    \item 2 image description pairs, each containing two texts
\end{itemize}
Given these, you are tasked with:
\begin{itemize}
    \item Selecting the overall “winner” for each pair
    \item Rating the qualities of each text
    \item Copying and pasting the best contributing sentence for each text
\end{itemize}

\centering
\includegraphics[width=0.5\linewidth, keepaspectratio]{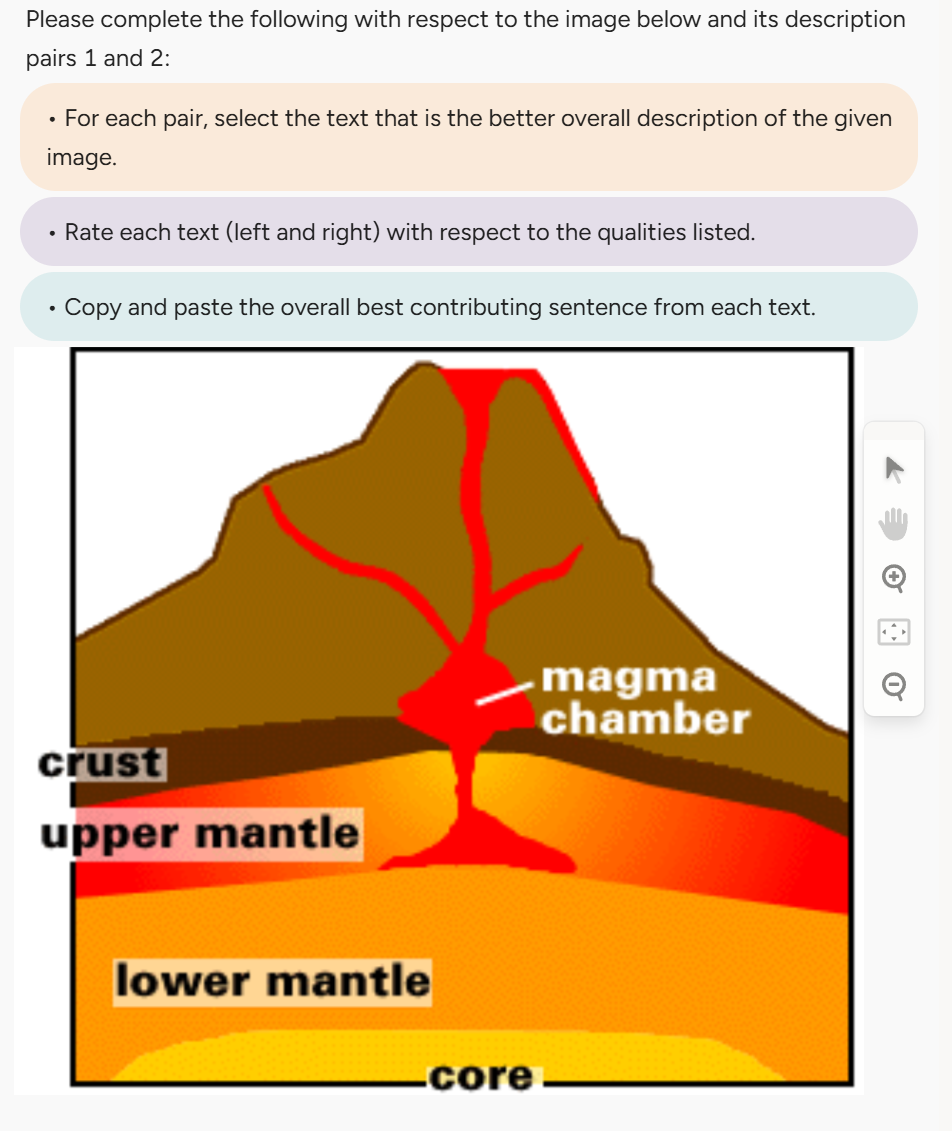} 
           
\end{tcolorbox}

\centering
\begin{tcolorbox}[
    colback=white,
    colframe=gray!20,
    arc=10pt,
    boxrule=1pt,
    title={\textcolor{white}{\textbf{Annotation Guidelines for the Sighted General Group (2/4)}}},
    coltitle=white,
    fonttitle=\large,
    colbacktitle=black!70,
]

Here is a detailed instruction on each task.

\paragraph{Selecting the overall winner}
\begin{itemize}
    \item Have a look at both texts. In your opinion, which is the better description of the text?
    \item This may be based on general impression or your internal criteria.
\end{itemize}

    \centering
    \includegraphics[width=0.95\linewidth]{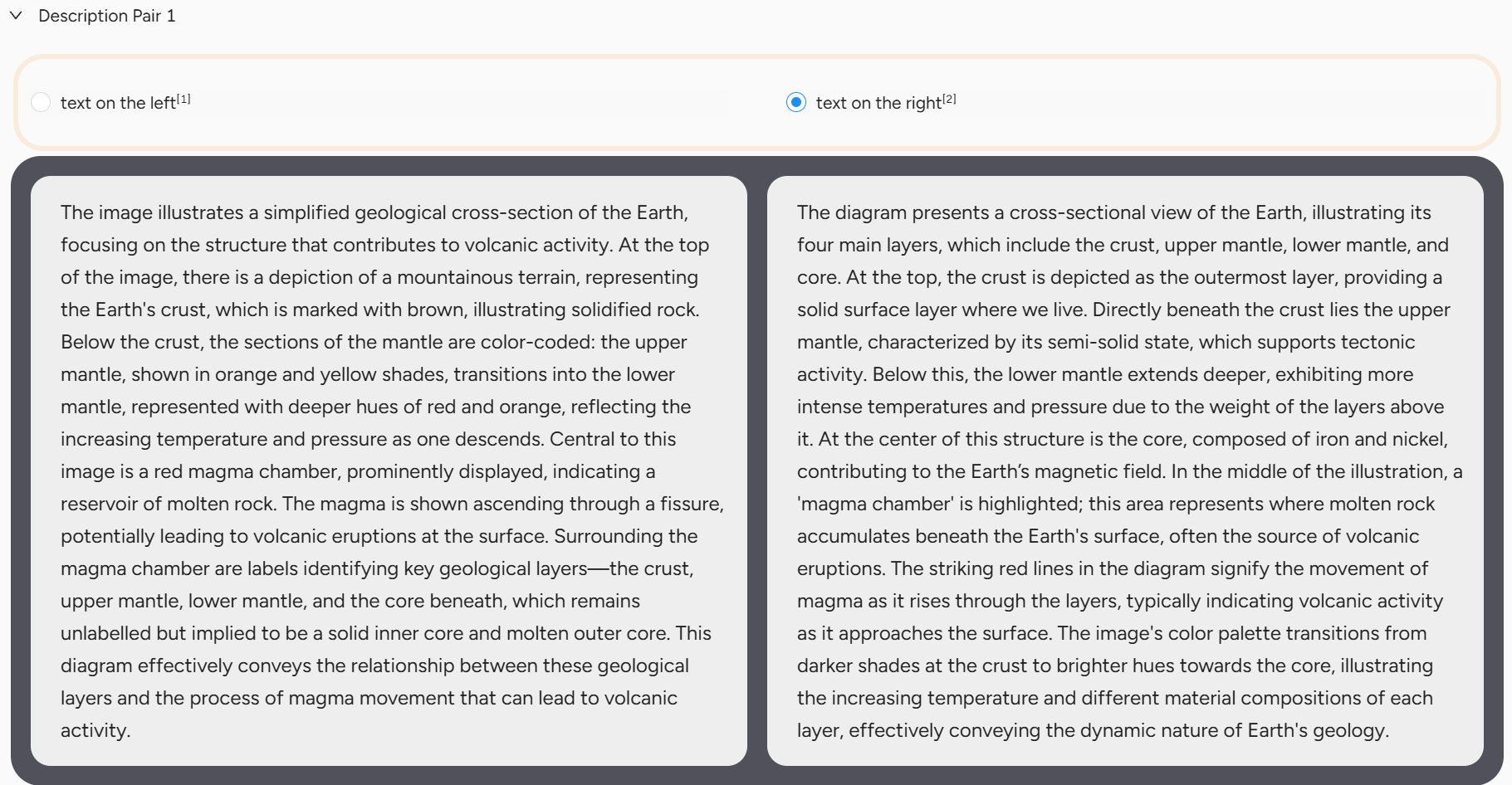}

\end{tcolorbox}

\centering
\begin{tcolorbox}[
    colback=white,
    colframe=gray!20,
    arc=10pt,
    boxrule=1pt,
    title={\textcolor{white}{\textbf{Annotation Guidelines for the Sighted General Group (3/4)}}},
    coltitle=white,
    fonttitle=\large,
    colbacktitle=black!70,
]

\paragraph{Rating the qualities of each text}
\begin{itemize}
    \item We break down what makes an image description a “good” one into a few qualities.
    \item Do not feel overly pressured to justify or rationalize your choice of the overall winner. In fact, you may treat the “overall winner” choice task completely independent of the quality rating task.
    \item On a scale from 1 to 5, how well does the text exhibit the following qualities?
    \begin{itemize}
        \item Factuality:
        \begin{itemize}
            \item Does the text contain facts about the image content?
            \item You may give a low score if the text contains statements that cannot be inferred from the image. These extraneous claims can include knowledge from the “world external to the image”, even though that knowledge itself may be true.
            \item You may give a low score if the text contains wrong statements about the image.
        \end{itemize}
        \item Informativeness:
        \begin{itemize}
            \item Does the text holistically describe the image content and help you become better informed about it?
            \item You may give a low score if some parts of the image seem “left out” in the text description.
        \end{itemize}
        \item Succinctness:
        \begin{itemize}
            \item Does the text describe the image content in a concise yet helpful way?
            \item Judgments based solely on text length should be avoided.
            \item Instead, think of the “density” of information contained in the text.
            \item You may give a low score if the text contains redundant/repeated information, inefficient sentence structures, and/or overly simple vocabulary that tend to make the text feel “sparse”.
        \end{itemize}
        \item Diversity:
        \begin{itemize}
            \item Does the text help you understand the image in various ways?
            \item There may be multiple effective descriptors about one image. There may be multiple perspectives and/or approaches to understanding one image. Do you think the given text addresses these?
            \item You may give a low score if the text feels too focused on small parts of the image or views the image in an overly specific (possibly contrived) perspective to lay out the description.
        \end{itemize}
    \end{itemize}
\end{itemize}

\begin{center}
    \includegraphics[width=0.95\linewidth]{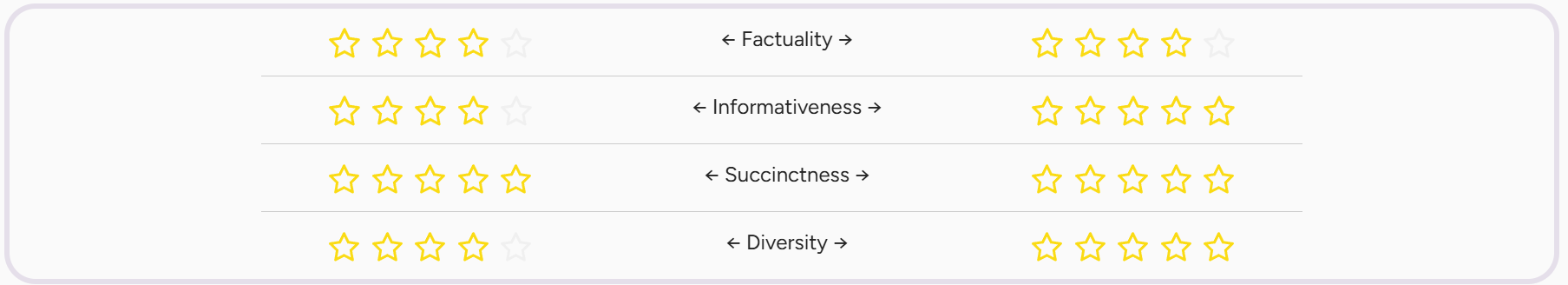}    
\end{center}

\end{tcolorbox}

\centering
\begin{tcolorbox}[
    colback=white,
    colframe=gray!20,
    arc=10pt,
    boxrule=1pt,
    title={\textcolor{white}{\textbf{Annotation Guidelines for the Sighted General Group (4/4)}}},
    coltitle=white,
    fonttitle=\large,
    colbacktitle=black!70,
]

\paragraph{Copying and pasting the best sentence for each text}
\begin{itemize}
    \item For each text, drag (highlight) the sentence that has best contributed to each text overall.
    \item The best sentence does not have to function as a one-sentence summary.
    \item This should be from the English text, not the Korean translation.
    \item Paste the sentence into the text field and press enter to submit.
    \item You will see your submitted best sentence on the display. Press the trash can button (Delete) to change your mind and submit a different sentence.
\end{itemize}

\begin{center}
    \includegraphics[width=0.95\linewidth]{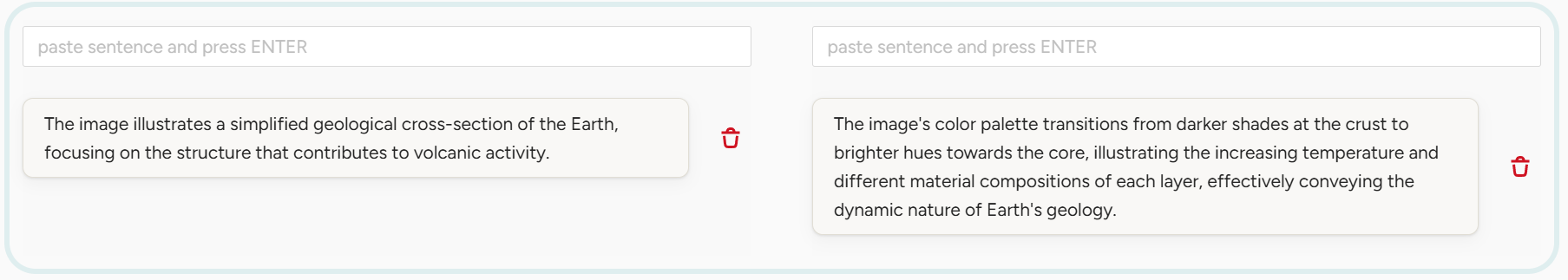}    
\end{center}

\paragraph{Caution}
\begin{itemize}
    \item Text length should not be a criterion for your assessment.
    \item Best sentence should be copied from the English paragraph.
\end{itemize}

\end{tcolorbox}

\centering
\begin{tcolorbox}[
    colback=white,
    colframe=gray!20,
    arc=10pt,
    boxrule=1pt,
    title={\textcolor{white}{\textbf{Evaluation Guidelines for the \textcolor{lime}{Sighted Educator} Group (1/2)}}},
    coltitle=white,
    fonttitle=\large,
    colbacktitle=black!70,
]

Thank you again for joining our experiments.

Attached is a spreadsheet containing 40 images, each of which has two descriptions, produced by various AI models in response to our request to generate a context for the input diagram.

As annotators, you are tasked with evaluating the quality of these texts.

\paragraph{Selecting the Preferred Text}

Have a look at the diagram in the ``Image'' column, along with the two contexts written in the ``Context1'' and ``Context2'' columns. In the ``Preferred Text'' column, enter your choice as 1 or 2.
\begin{itemize}
    \item This preference may rely on your own personal criteria.
\end{itemize}

\end{tcolorbox}

\centering
\begin{tcolorbox}[
    colback=white,
    colframe=gray!20,
    arc=10pt,
    boxrule=1pt,
    title={\textcolor{white}{\textbf{Evaluation Guidelines for the \textcolor{lime}{Sighted Educator} Group (2/2)}}},
    coltitle=white,
    fonttitle=\large,
    colbacktitle=black!70,
]

\paragraph{Quantitative Assessment}

Give a score ranging from 1 to 5 for each quality listed below. For each statement, enter your assessment on a scale from a 5 if you ``Strongly Agree'' to a 1 if you ``Strongly Disagree''.

\begin{itemize}
    \item Factuality: The text delivers only facts that are grounded on the diagram.
    \begin{itemize}
        \item Even if a piece of knowledge in the text is factual, you may give a low score if that knowledge cannot be inferred from the diagram.
        \item Wrong textual descriptions of the diagram content also have low merit.
    \end{itemize}
    \item Informativeness: The text describes all of the diagram, holistically.
    \begin{itemize}
        \item You may give a low score to a context leaving out parts of the diagram.
    \end{itemize}
    \item Succinctness: The text is concise and to the point.
    \begin{itemize}
        \item Please assess whether the context conveys an appropriate ``density'' of information.
        \item You may give a low score if a context seems repetitive.
        \item Please avoid scoring based on apparent text length.
    \end{itemize}
    \item Diversity: The text captures a variety of perspectives from the diagram and employs multiple effective ways of getting the diagram message across.
    \begin{itemize}
        \item There may be multiple different ways to understand a diagram. Please assess whether these ways have been put together in the given context.
    \end{itemize}
    \item Usefulness: The text is helpful to BLV.
    \begin{itemize}
        \item As an experienced educator for learners with visual impairments, please evaluate how useful and helpful the text would be.
    \end{itemize}
\end{itemize}

\paragraph{Qualitative Assessment}

In the ``Reason'' column, please justify your preference choice (\textit{i.e.}, the 1 or 2 selection) with a brief explanation. Simple comments, as long as they tell us the textual quality your choice was based on, may still prove helpful for our research. See examples below.

\begin{itemize}
    \item ``contains various descriptions of ants''
    \item ``written more logically''
    \item ``Context 1 contains a more realistic definition of a food web.''
    \item ``2 is more concise.''
    \item ``Context 1 lacks a description of the artery.''
    \item ``While both texts faithfully address the rotation and revolution movements of the Earth, Context 1 describes in-depth how they manifest as different natural phenomena on the planet.''
\end{itemize}

\end{tcolorbox}

\centering
\begin{tcolorbox}[
    colback=white,
    colframe=gray!20,
    arc=10pt,
    boxrule=1pt,
    title={\textcolor{white}{\textbf{Evaluation Guidelines for the \textcolor{Yellow}{BLV Educator} Group}}},
    coltitle=white,
    fonttitle=\large,
    colbacktitle=black!70,
]

Thank you again for joining our experiments.

Attached is a spreadsheet containing 80 images, each with a description that has been produced by various AI models in response to our request to generate a context for the input diagram.\\

The spreadsheet comes with 81 rows and 8 columns. The table headers are as follows: Image, Context, Succinctness, Diversity, Usefulness (Summary), Usefulness (Multiple-choice Questions), Usefulness (Open-ended Questions), and Nature of Context. Apart from the header row, the remaining 80 rows each contain 1 image and 1 description.\\

As annotators, you are tasked with evaluating these texts, based on the text alone.

\paragraph{Quantitative Assessment}

Give a score ranging from 1 to 5 for each quality listed below. For each statement, on a scale from a 5 if you ``Strongly Agree'' to a 1 if you ``Strongly Disagree''.

\begin{itemize}
    \item Succinctness: The text is concise and to the point.
    \begin{itemize}
        \item Please assess whether the context conveys an appropriate ``density'' of information.
        \item You may give a low score if a context seems repetitive.
        \item Please avoid scoring based on apparent text length.
    \end{itemize}
    \item Diversity: The text captures a variety of perspectives from the diagram and employs multiple effective ways of getting the diagram message across.
    \begin{itemize}
        \item There may be multiple different ways to understand a diagram. Please assess whether these ways have been put together in the given context.
    \end{itemize}
    \item Usefulness (Summary): The text serves as a good summary.
    \begin{itemize}
        \item Please assess how well the text helps you formulate an idea of the diagram content.
    \end{itemize}
    \item Usefulness (Multiple-choice Questions): The text would be useful in solving short-answer, multiple-choice questions based on the diagram.
    \begin{itemize}
        \item Suppose you are to solve short-answer multiple-choice questions that have been constructed from the diagram. How well would the context help you answer these questions?
    \end{itemize}
    \item Usefulness (Open-ended Questions): The text would be useful in solving descriptive, essay type questions based on the diagram.
    \begin{itemize}
        \item Suppose you are to answer an open-ended long-answer question that has been constructed from the diagram. How well would the context help you answer such a question?
    \end{itemize}
    \item Nature of Context: The text is rich in interpretive detail.
    \begin{itemize}
        \item On a scale of 1 to 5, if the text appears to lay out plain and straightforward facts from the diagram, give a score of 1. If it rather contains interpretive descriptions, and/or reasoned explanations, give a score of 5.
    \end{itemize}
\end{itemize}

\end{tcolorbox}

\end{document}